\documentclass[twocolumn]{IEEEtran}
\usepackage{amsmath,amsfonts}
\usepackage{algorithmic}
\usepackage{algorithm}
\usepackage{array}
\usepackage[caption=false,font=normalsize,labelfont=sf,textfont=sf]{subfig}
\usepackage{textcomp}
\usepackage{stfloats}
\usepackage{url}
\usepackage{verbatim}
\usepackage{graphicx}
\usepackage{cite}

\def\BibTeX{{\rm B\kern-.05em{\sc i\kern-.025em b}\kern-.08em
    T\kern-.1667em\lower.7ex\hbox{E}\kern-.125emX}}
\usepackage{balance}

\usepackage{microtype}
\usepackage[utf8]{inputenc} 
\usepackage[T1]{fontenc}    
\usepackage{paralist} 
\usepackage{marvosym} 
\usepackage{bm} 
\usepackage{bbm} 
\usepackage{amssymb}   
\usepackage{mathrsfs} 
\usepackage{tabularray} 
\UseTblrLibrary{booktabs}       
\usepackage{tabularx} 
\usepackage[flushleft]{threeparttable} 
\usepackage{colortbl} 
\usepackage{makecell} 
\usepackage{multirow} 
\usepackage{nicefrac} 
\usepackage[dvipsnames]{xcolor} 
\usepackage{soul} 
\setul{0.5ex}{0.3ex}

\definecolor{citecolor}{RGB}{34,139,34}
\usepackage[pagebackref=true,breaklinks=true,letterpaper=true,colorlinks,
    citecolor=citecolor,bookmarks=false]{hyperref}
\usepackage{cleveref}  
\crefformat{figure}{Fig.~#2#1#3}
\crefformat{equation}{Eq.~(#2#1#3)}
\crefformat{table}{Table~#2#1#3}
\crefformat{section}{Section~#2#1#3}
\crefformat{algorithm}{Algorithm~#2#1#3}
\crefformat{appendix}{Appendix #2#1#3}
\crefformat{subsection}{subsection~#2#1#3}
\usepackage{cellspace}

\usepackage{xspace} 

\usepackage{pifont} 
\definecolor{modifiedorange}{RGB}{255,140,0} 

\usepackage{float}
\usepackage{placeins}

\definecolor{Gray}{rgb}{0.9,0.9,0.9}
\definecolor{LightCyan}{rgb}{0.88,1,1}
\newcolumntype{a}{>{\columncolor{Gray}}c}
\newcolumntype{b}{>{\columncolor{white}}c}

\begin{document}
\setlength{\abovedisplayskip}{.5\baselineskip} 
\setlength{\belowdisplayskip}{.5\baselineskip} 

\title{Beyond Unfolding: 60$\times$ Faster One-Stage Unmixing for Closely-Spaced Infrared Small Targets}



\author{
  Ximeng~Zhai,
  Zheng~Wang,
  Yaohong~Chen,
  Hao~Wang,
  Ming-Ming Cheng,
  Yimian~Dai
  \thanks{
    This work was supported by the National Science Foundation of China (No. 62301261, No. 62225604), the Tianjin Natural Science Foundation Project (No. 25JCQNJC01370), 
    the West Light Foundation of the Chinese Academy of Sciences (XAB2022YN06),
    the Shenzhen Science and Technology Program (No. JCYJ20240813114237048), the Fundamental Research Funds for the Central Universities (No. 63253217, No. 63253217), and the Supercomputing Center of Nankai University (NKSC).
    \emph{
    (Corresponding author: Yimian Dai).}
    }

  \thanks{
    Ximeng Zhai, Yaohong Chen, and Hao Wang are with Xi'an Institute of Optics and Precision Mechanics, Chinese Academy of Sciences, Xi'an 710119, China. Zhai is also with the University of Chinese Academy of Sciences, Beijing 100049, China
    (e-mail:
    \href{mailto:zhaiximeng23@mails.ucas.ac.cn}{zhaiximeng23@mails.ucas.ac.cn}).    
  }
  
  \thanks{
  Zheng~Wang is with Beijing Institute of Astronautical Systems Engineering, Beijing 100094, China
  (e-mail:
  \href{wangzhengjack@163.com}{wangzhengjack@163.com}).
  }
  \thanks{
  Ming-Ming Cheng and Yimian Dai are with VCIP, College of Computer Science, Nankai University, Tianjin 300071, China. They are also affiliated with the NKIARI, Shenzhen Futian, Shenzhen 518045, China (e-mail:
  \href{mailto:cmm@nankai.edu.cn}{cmm@nankai.edu.cn};
  \href{mailto:yimian.dai@gmail.com}{yimian.dai@gmail.com}).
  }
}


\maketitle

\begin{abstract}
Due to the optical diffraction limit and long imaging distances, Closely-Spaced Infrared Small Targets (CSIST) typically exhibit energy overlap, manifesting as indistinguishable blobs in infrared images. 
This ambiguity invalidates the one-to-one mapping assumption of traditional detection, thereby necessitating a paradigm shift towards CSIST Unmixing, which decomposes these blobs into discrete sub-targets. 
However, the dominant paradigm deep unfolding networks (DUNs) are shackled by the high latency and structural inflexibility intrinsic to their repetitively iterative architecture.
To this end, we propose the \textbf{Fast One-stage CSIST Unmixing Scheme (FOCUS)}, a one-stage lightweight paradigm which demonstrates that deep unfolding is not necessary for CSIST Unmixing.
Motivated by the key observation that image super-resolution (SR) and CSIST Unmixing share an isomorphic degradation model, our insight is that it is possible to achieve a paradigm shift from image SR to CSIST Unmixing via completely transforming the label space, loss functions, and evaluation criteria.
Specifically, to avoid entangling geometric recovery with artifact suppression, FOCUS adopts a single pass mapping with an internal coarse-to-fine flow that progressively refines target localization from coarse spatial distributions to finer sub-pixel precision.
While sparsity regularization suppresses background clutter, it also attenuates target intensities.
To compensate for this attenuation of valid signals, flux conservation is introduced as a competing constraint that restores signal energy back to target centers.
To the best of our knowledge, this work is the first attempt to address the CSIST Unmixing task via a lightweight one-stage framework without the DUN paradigm. 
Experiments demonstrate that our method matches or surpasses the state-of-the-art unfolding approaches in both localization and unmixing accuracy, while boosting the inference speed by \textbf{60$\times$}. 
Code is available at \url{https://github.com/GrokCV/GrokCSO}.
\end{abstract}

\begin{IEEEkeywords}
Closely-spaced infrared small target unmixing,
one-stage lightweight inference,
flux conservation,
deep learning
\end{IEEEkeywords}





\section{Introduction} \label{sec:introduction}

\begin{figure}[h]
    \flushright 
    \includegraphics[width=.48\textwidth]{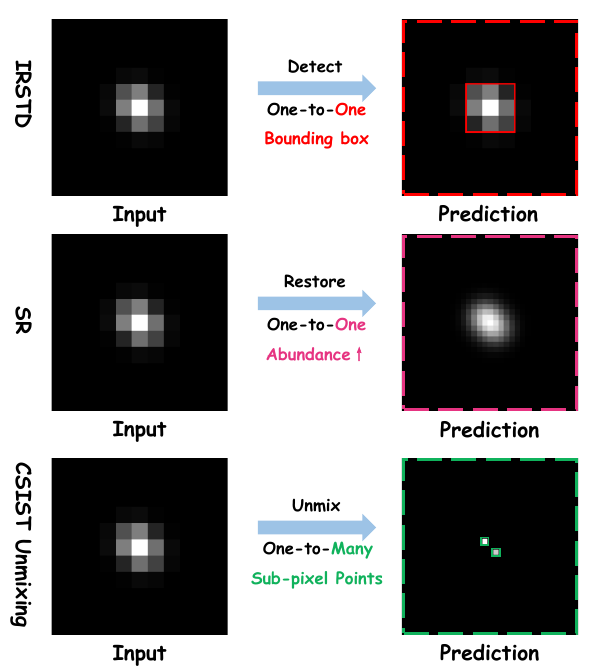}
    \caption{Conceptual comparison of IRSTD, SR, and CSIST Unmixing. IRSTD utilizes a one-to-one paradigm to identify targets through bounding boxes. SR focuses on visual restoration while still maintaining a one-to-one correspondence. In contrast, CSIST Unmixing adopts a one-to-many paradigm to recover discrete sub-pixel point sources from overlapped energy regions.}
    \label{fig:IRSTD_VS_SR_VS_CSISTU}
\end{figure}

\IEEEPARstart{I}{nfrared} imaging plays an indispensable role in critical fields such as early warning systems, maritime surveillance, and border protection, owing to its robust capability to operate under low-light conditions and penetrate atmospheric obscurants \cite{zhang2025wmrnet}.  
However, with the increasing demand for long-distance detection \cite{xiong2025drpca}, targets of interest, such as distant aircraft, often manifest as dim spots with faint intensity and low contrast due to the long imaging distance \cite{chen2024sstnet}.
This imposes stringent requirements on the accuracy and robustness of Infrared Small Target Detection (IRSTD) systems.

Nevertheless, \textbf{current IRSTD research mainly focuses on target localization (i.e., ``where is the target?'').}
Existing research has spanned from classical filtering \cite{chen2013local} and decomposition \cite{feng2023coarse} to deep learning architectures \cite{wu2026neural}, where the core attempts are to pop out the dim targets while suppressing false alarms.
This status quo (task formulation) is restricted by a default assumption of \textbf{strict one-to-one mapping} between diffraction spots and targets, where each spot in the image corresponds to exactly one real-world target \cite{zhai2025seqcsist}.

However, in practice, as the targets become dense, such as drone swarms, and imaging distance becomes larger, \textit{the optical energy diffusion causes multiple adjacent targets to merge into a single overlapping blob} \cite{macumber2005hierarchical}, 
which leads to the so-called \textit{Closely-Spaced Infrared Small Target (CSIST)} scenarios.
Traditional IRSTD methods suffer from missed counts and lost coordinates in these scenarios because of the one-to-one assumption, as illustrated in Fig.~\ref{fig:IRSTD_VS_SR_VS_CSISTU}.
To surpass this constraint, research has transitioned from coarse detection toward precise unmixing, which constitutes \textit{the specialized task termed CSIST Unmixing} \cite{zhai2025seqcsist}.
The task is to disentangle energy-overlapped blobs into discrete point sources at the sub-pixel level, thereby recovering their precise coordinates and intensities. 
\textbf{Therefore, very different from target detection, CSIST Unmixing is in essence a one-to-many inverse mapping problem. }

Current CSIST Unmixing schemes rely on deep unfolding paradigm by mapping iterative algorithms into unrolled layers.
For example, DISTA-Net \cite{han2025dista} unrolled iterative shrinkage thresholding into a dynamic framework.
DeRefNet \cite{zhai2025seqcsist} extended this paradigm to sequential data by employing inter-frame alignment.
These two methods laid the foundation for single-frame and multi-frame CSIST Unmixing.
Recently, STSA-Net \cite{xiao2026stsa} refined this paradigm by incorporating target sensitive convolutions and pixel wise adaptive thresholds within an unfolding structure. 
Later, DSCSNet \cite{tang2026dscsnet} utilized an alternating direction method of multipliers (ADMM) \cite{neal2011distributed} based unrolled architecture with strict sparsity constraints to protect radiant energy peaks.
These models collectively demonstrate that the deep unfolding paradigm provides an effective approach for resolving CSIST scenarios.

Despite their effectiveness, the reliance on unrolled iterations imposes structural constraints \cite{he2026unfoldir}. 
Recursive applications of operators result in high computational overhead, which leads to inference delays. 
Moreover, the coupling between network layers and mathematical steps limits the flexibility for modular design or lightweight optimization. 
These drawbacks necessitate a direct mapping architecture capable of high speed inference without the computational overhead of iterative loops.

In this paper, we propose a paradigm shift, which demonstrates that \textbf{unfolding is not necessary for CSIST Unmixing.}
Indeed, while unfolding provides a path for unmixing, it represents one possible choice rather than a unique solution.
This stance originates from our key observation that CSIST Unmixing and image super-resolution (SR) share a common mathematical basis, as both tasks reconstruct the original scene from the isomorphic degradation model ($Y = Hx + n$).
Accordingly, a natural question araises, \textit{can we discard the cumbersome DUNs paradigm and instead leverage the highly efficient, lightweight end-to-end networks from the SR field to solve the CSIST Unmixing problem?}

Our answer is affirmative.
By transforming the label space, loss functions, and evaluation criteria, 
we successfully achieve a paradigm shift, wherein the image SR paradigm is modified to resolve the CSIST Unmixing problem.
This transition redirects the optimization of generic convolutional networks from reconstructing visual textures toward the precise inversion of discrete point sources.
In addition, the strategy eliminates the dependency on iterative unfolding structures, which ensures architectural flexibility while improving inference speed.

To this end, we propose FOCUS (Fast One-stage CSIST Unmixing Scheme), a lightweight one-stage architecture, which consists of two dedicated modules underpinned by a shared isomorphic unit.
Distinct from generic image SR networks that conflate spatial reconstruction with artifact suppression, FOCUS adopts a coarse-to-fine internal flow that explicitly separates the two objectives: the first module recovers spatial topology from low-resolution observations, while the second performs residual refinement to suppress upsampling artifacts at full resolution.
To concentrate signal energy into precise target locations, a dual-branch gated structure with sparsity regularization suppresses background responses while preserving target centroids at sub-pixel resolution.
However, sparsity enforcement inevitably attenuates target signal amplitudes alongside background suppression. Flux conservation is therefore introduced as a competing constraint that anchors total reconstructed energy to its physical ground truth, ensuring that suppressed energy is restored to target centers rather than lost.
Experiments demonstrate that FOCUS achieves a 60$\times$ speedup over iterative baselines while maintaining competitive unmixing precision.

Our main contributions are summarized as follows:
\begin{enumerate}
    \item \textbf{Novel Paradigm:} We reformulate CSIST Unmixing as one-stage mapping, replacing iterative DUNs with lightweight feed-forward inference.
    \item \textbf{Constrained Design:} We propose FOCUS with sparsity regularization and flux conservation constraints, jointly mitigating background interference and amplitude attenuation through balanced constraint trade-offs.
    \item \textbf{Efficiency Improvement:} FOCUS achieves a 60$\times$ speedup over iterative paradigms while maintaining competitive unmixing accuracy.
\end{enumerate}



\section{Related Work} \label{sec:related}

\subsection{Image Super-Resolution}
As a fundamental low-level vision task \cite{pan2018learning}, image SR aims to reconstruct high-fidelity details from degraded observations, where the lost high-frequency information recovers from incomplete measurements \cite{wen2026incorporating}.

Since the pioneering work of SRCNN \cite{dong2015image_SRCNN}, which established the paradigm of end-to-end convolutional mapping, the field has witnessed a rapid evolution in architectural design. 
Subsequent research has largely bifurcated into two directions: maximizing reconstruction quality through deeper residual networks (e.g., SAN \cite{dai2019second}, SRResNet \cite{ledig2017photo}) and optimizing inference efficiency via lightweight structure (e.g., IMDN \cite{Hui-IMDN-2019}, RFDN \cite{liu2020residual}). 
These advances have endowed modern SR networks with powerful non-linear fitting capabilities and efficient feature extraction mechanisms.

Nevertheless, as shown in Fig.~\ref{fig:IRSTD_VS_SR_VS_CSISTU}, while image SR aims to overcome resolution limits, its primary objective is \textit{visual restoration} \cite{mou2024empowering}, enhancing textural richness to generate continuous and aesthetically pleasing images.
If a blurred infrared blob is fed into a standard SR system, the output remains a refined blob with smoother edges and higher granularity, rather than physically separated point sources as CSIST Unmixing does \cite{blau2018perception}.

Driven by these divergent objectives, the technical trajectories of SR and CSIST Unmixing have bifurcated.
The SR community has largely embraced purely data-driven convolutional neural networks (CNNs) (e.g., ESPCN \cite{shi2016real_ESPCN}, DBPN \cite{haris2018deep_DBPN}).
In contrast, the CSIST Unmixing community predominantly favors model-driven deep learning, particularly DUNs (e.g., DISTA-Net \cite{han2025dista}, DeRefNet \cite{zhai2025seqcsist}).
These methods unroll iterative optimization algorithms into neural layers, embedding sparse priors to solve the imaging inverse problem.

Our work bridges this gap through the following two highlights:

\begin{enumerate}
    \item \textbf{Cross Field Adaptation:} Unlike conventional image SR models limited to low level visual restoration, this scheme repurposes SR mapping for high level CSIST Unmixing to recover discrete point objects.
    \item \textbf{Flux-Preserving Sparse Strategy:} Compared to generic image SR methods that rely solely on smoothness optimization, this work pioneers the integration of structural sparsity and flux conservation as competing training constraints, achieving a balance between source sharpness and amplitude fidelity.
\end{enumerate}

\subsection{Closely-Spaced Infrared Small Target Unmixing}
Over the past decade, Model-Driven Deep Learning, specifically DUNs, has established itself as the dominant paradigm for solving inverse problems in computational imaging \cite{monga2021algorithm}. 
By unrolling iterative optimization algorithms (e.g., ISTA \cite{daubechies2004iterative}, ADMM \cite{neal2011distributed}) into cascade neural architectures, DUNs integrate the interpretability of physical models with the learning capability of deep networks \cite{zhang2018istanet_ISTANet, yang2018admm}.

In the specific domain of CSIST Unmixing, pioneering works like DISTA-Net \cite{han2025dista} and DeRefNet \cite{zhai2025seqcsist} have crystallized a rigorous paradigm: reformulating the unmixing task as a sparsity-driven signal recovery problem within a deep learning framework. 
Subsequently, this paradigm is extended by STSA-Net \cite{xiao2026stsa} with target-sensitive proximal operators and by DSCSNet \cite{tang2026dscsnet} via an ADMM-based unrolling with strict $\ell_1$-norm sparsity constraints.
Fundamentally, these methods explicitly construct the sensing matrix based on the Point Spread Function (PSF) \cite{nayar2004motion}.
By unrolling the alternation between data fidelity and sparsity regularization into learnable network stages, they effectively guide the latent features to migrate from the diffuse observation domain towards a high-dimensional discrete solution space. 
This mechanism allows for interpretable source disentanglement, and has demonstrated the potential to overcome the diffraction limit.

Although these CSIST Unmixing methods provide theoretical guarantees, they share a common limitation: they are constrained by the rigidity of the physical operators and the recurrent nature of iterative structure. 
The explicit reliance on fixed degradation operators makes it difficult to adapt to spatially variant energy distributions, while the depth of the network directly correlates with inference latency.
Our work addresses these structural bottlenecks through the following two highlights:
\begin{enumerate}
    \item \textbf{Paradigm Shift:} We transition from the rigid, multi-stage iterative architecture of DUNs to a flexible, single-stage feed-forward paradigm. This fundamental shift eliminates recurrent optimization bottlenecks, yielding a streamlined architecture that is significantly easier to optimize and adapt.
    \item \textbf{Inference Acceleration:} By replacing heavy recurrent operators with a streamlined one-stage feed-forward pipeline, we achieve a drastic reduction in computational burden.
This approach dramatically reduces computational latency without sacrificing sub-pixel localization accuracy.
\end{enumerate}

\section{Revisiting CSIST Unmixing: An Isomorphic Perspective from Image Super-Resolution} \label{sec:revisiting}

\begin{figure}[h]
    \flushright 
    \includegraphics[width=.48\textwidth]{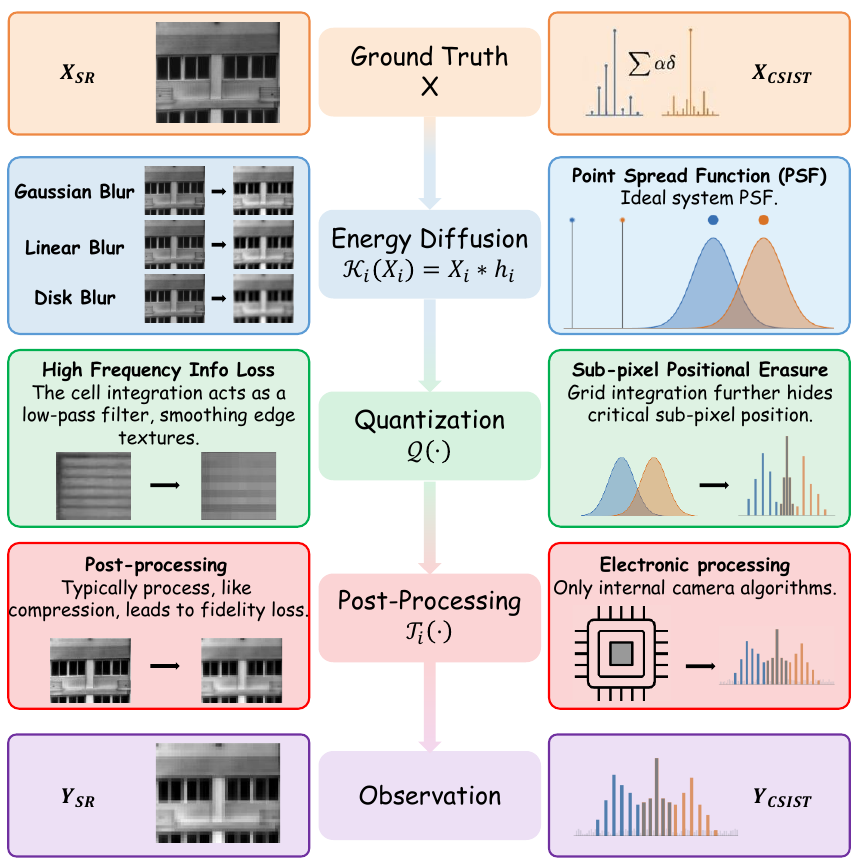}
    \caption{A visual comparison of the isomorphic forward imaging pipelines for image Super-Resolution (left) and CSIST unmixing (right). Both tasks share a unified mathematical structure composed of energy diffusion ($\mathcal{K}$), spatial quantization ($\mathcal{Q}$), and post-processing ($\mathcal{T}$), while diverging in their physical priors and latent signal characteristics.}
    \label{fig:CSISTU_SR}
\end{figure}

Both image SR and CSIST Unmixing seek to reconstruct a high-resolution signal $\mathbf{X}$ from a degraded low-resolution observation $\mathbf{Y}$. 
Despite differing application contexts, the imaging processes of two tasks share a unified mathematical framework \cite{yang2010image}. 

As illustrated in Fig.~\ref{fig:CSISTU_SR}, the forward imaging processes of these two tasks follow a parallel structure.
Let the subscript $i \in \{\text{SR}, \text{CSIST}\}$ denote the specific task. 
The generalized observation model is formulated as:
\begin{equation}
    \mathbf{Y}_i = \mathcal{D}_i(\mathbf{X}_i) + \mathbf{n}_i
\end{equation}
where $\mathbf{n}_i \in \mathbb{R}^{h \times w}$ represents the additive system noise and $\mathcal{D}_i: \mathbb{R}^{H \times W} \to \mathbb{R}^{h \times w}$ is the composite degradation operator. Based on imaging system theory \cite{zhang2021designing}, the operator $\mathcal{D}_i$ can be decomposed into a sequential chain of isomorphic transformations:
\begin{equation}
    \mathcal{D}_i = \mathcal{T}_i \circ \mathcal{Q}_i \circ \mathcal{K}_i.
\end{equation}
While the structural mapping remains identical across both domains, the physical interpretation and the resulting latent priors diverge during the instantiation of each operator.

\subsubsection{\textbf{Optical Blur ($\mathcal{K}$)}}
The operator $\mathcal{K}_i$ models the spatial dispersion of radiant energy. 
This process is generalized as a convolution between the ground truth signal and a specific kernel:
\begin{equation}
    \mathcal{K}_i(\mathbf{X}_i) = \mathbf{X}_i \ast h_i.
\end{equation}
In Image SR, the kernel $h_{\text{SR}}$ is generally artificially simulated to mimic resolution loss, employing mathematical proxies such as Gaussian kernels for isotropic blur, directional linear kernels for motion blur, or disk kernels for defocus. 
In contrast, for CSIST unmixing, the degradation $h_{\text{CSIST}}$ is a physical PSF inherent to the optical assembly. 
It encapsulates complex wave-optical artifacts including wavefront aberrations, coma, and field-dependent distortions that perturb the sub-pixel energy distribution.

\subsubsection{\textbf{Quantization ($\mathcal{Q}$)}}
The operator $\mathcal{Q}_i$ describes the mapping from the continuous optical domain to the discrete digital domain. Both tasks involve grid level integration over the detector photosensitive area $\Omega$:
\begin{equation}
    \mathcal{Q}_i(\cdot)[j, k] = \iint_{\Omega_{j,k}} (\cdot) \, dx dy.
\end{equation}
While both tasks suffer from the loss of information during this sampling, the implications differ. 
In SR, this discretization filters out high-frequency textural details. 
In CSIST Unmixing, the integration of overlapped energy distributions into pixel intensities further destroys the sub-pixel positional information. 
This shared degradation structure renders both tasks a highly underdetermined inverse problem.

\subsubsection{\textbf{Post-processing ($\mathcal{T}$)}}
For SR, $\mathcal{T}_{\text{SR}}$ often involves algorithmic level processing such as compression for storage efficiency or augment training robustness, which usually leads to digital smoothing, contrast adjustment, or downsampling. 
For unmixing, $\mathcal{T}_{\text{CSIST}}$ is primarily restricted to electronic-level degradation, reflecting the readout circuitry's impact on the infrared signal before it reaches the digital interface.

\subsubsection{\textbf{Additive System Noise ($\mathbf{n}$)}}
The noise term $\mathbf{n}$ further differentiates the tasks. In SR pipelines, $\mathbf{n}_{\text{SR}}$ frequently includes both natural sensor noise and artificially injected noise (e.g., JPEG artifacts or synthetic Gaussian noise) to improve model generalization. Conversely, $\mathbf{n}_{\text{CSIST}}$ is strictly modeled based on natural stochastic processes within the infrared detector, such as dark current and thermal fluctuations.

\subsubsection{\textbf{Theoretical Basis for Paradigm Transfer}}
The structural isomorphism between $\mathcal{D}_{\text{SR}}$ and $\mathcal{D}_{\text{CSIST}}$ suggests that the feature extraction architectures perfected for image restoration can be repurposed for physical unmixing. By respecting the underlying sparsity and radiant flux conservation, the paradigm can shift from visual aesthetics toward the precise parameter inversion of discrete radiant sources.

\section{FOCUS: Fast One-Stage CSIST Unmixing Scheme}

\begin{figure*}[!t]
    \centering
    \includegraphics[width=.98\textwidth]{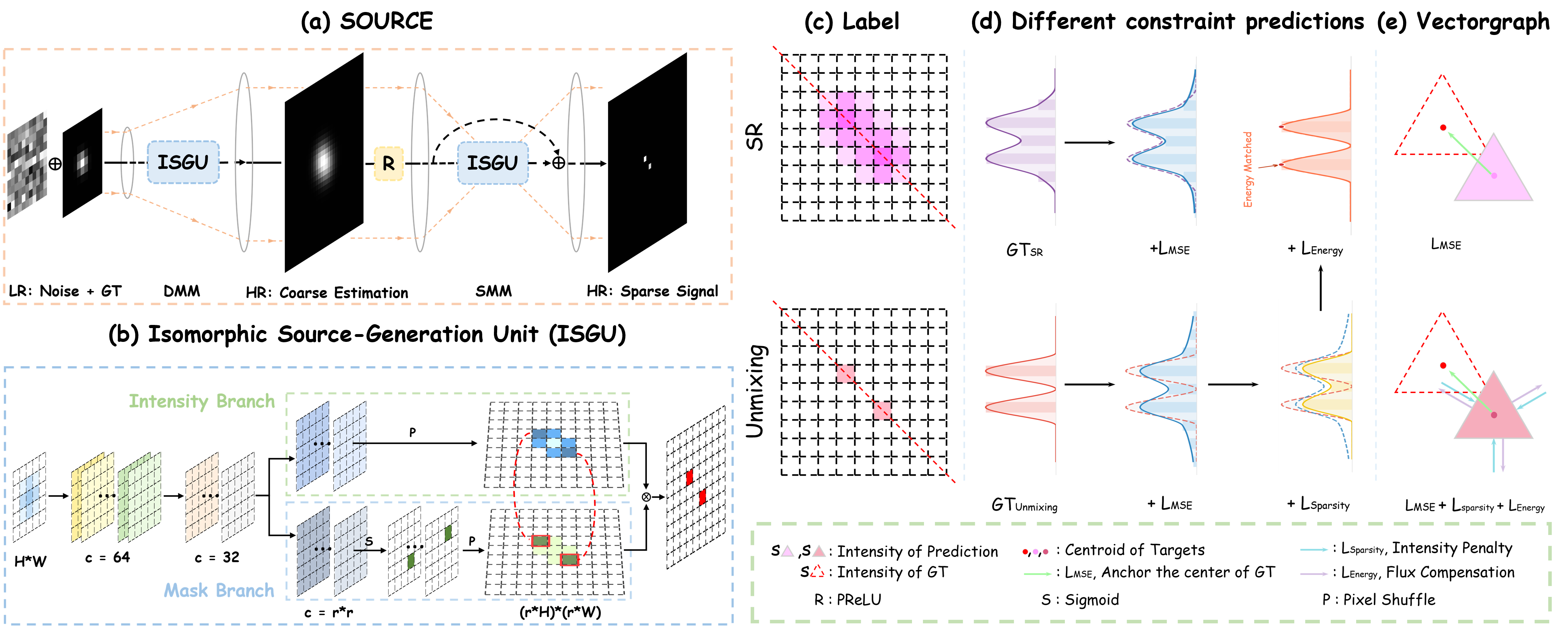}
    \caption{Overview of the FOCUS framework and its physics-constrained optimization strategy. (a-b) The architecture utilizes a coarse-to-fine flow via DMM and SMM modules, incorporating Isomorphic Source-Generation Units (ISGU) that bifurcate intensity and mask paths to ensure discrete source recovery. (c-e) Comparison of label characteristics and predictive evolution between standard super-resolution (SR) and the proposed unmixing paradigm. The evolution under joint $L_{\text{MSE}}$, $L_{\text{Sparsity}}$, and $L_{\text{Energy}}$ constraints, visualized through the optimization vectorgraph, illustrates the competitive equilibrium achieved to maintain structural sparsity and flux conservation for high-fidelity signal reconstruction.}
    \label{fig:FOCUS}
\end{figure*}

\subsection{\textbf{Overall Framework: A Progressive One-Stage Paradigm}}
Given the degradation model $\mathbf{y} = \Phi(\mathbf{x}) + \mathbf{n}$, the goal of CSIST Unmixing is to reconstruct the high-resolution sparse source map $\mathbf{x} \in \mathbb{R}^{rH \times rW}$ from the observed low-resolution input $\mathbf{y} \in \mathbb{R}^{H \times W}$, where $r$ is the upscaling factor. 
This constitutes a severe ill-posed inverse problem \cite{bertero2021introduction}.

To solve this efficiently, we propose the \textbf{FOCUS} framework. 
As illustrated in Fig.~\ref{fig:FOCUS} (a), FOCUS is structured as a One-Stage Progressive Coarse-to-Fine Flow. 
It is crucial to distinguish our one-stage paradigm from traditional one-stage networks (e.g., standard SRCNN \cite{dong2015image_SRCNN} or RCAN \cite{zhang2018image}).
Distinct from conventional black box paradigms where upsampling and restoration are inextricably entangled, FOCUS adopts a stratified strategy to explicitly decouple geometric recovery from physical fidelity rectification, thereby reducing optimization complexity to achieve rigorous physical precision within a unified feed-forward flow.
Thus, implementing the Decoupled Reconstruction-Rectification (DRR) strategy, we formulate the overall mapping $\mathcal{F}_{\Theta}$ as a global learning process.

Firstly, the input is projected into the high-resolution space via the \textit{Digital Microscope Module} ($\mathcal{G}_{\text{DMM}}$) to obtain a coarse geometric estimation. Subsequently, to suppress the upsampling artifacts inherent in this estimation, the \textit{Spatial Modulator Module} ($\mathcal{R}_{\text{SMM}}$) acts as a residual learner. It predicts a sparse correction map, which is then superimposed onto the coarse estimation via a global skip connection. Mathematically, we encapsulate this progressive coarse-to-fine evolution into a unified mapping function:
\begin{equation}
    \hat{\mathbf{x}} = \mathcal{F}_{\Theta}(\mathbf{y}) = \underbrace{\mathcal{R}_{\text{SMM}} \left( \underbrace{\sigma(\mathcal{G}_{\text{DMM}}(\mathbf{y}))}_{\text{Coarse Estimation}} \right)}_{\text{Fine-grained Rectification}}
    \label{eq:overall_framework}
\end{equation}
where ${F}_{\Theta}$ denotes the overall mapping function parameterized by ${\Theta}$. $\sigma(\cdot)$ represents the PReLU \cite{he2015delving} activation, which serves to rectify the geometric estimation into a valid non-negative feature space before residual learning.

The flow consists of two physically distinct phases:

\textbf{Digital Microscope Module (DMM):} Analogous to an optical microscope that magnifies tiny specimens, this module ($\mathcal{G}_{\text{DMM}}$) acts as a geometric projector. It is responsible for the inverse projection of the signal from the low-dimensional observation space to the high-dimensional target space ($11 \times 11 \to 33 \times 33$), establishing the spatial topology of point sources.

\textbf{Spatial Modulator Module (SMM):} Analogous to a spatial light modulator that shapes wavefronts without changing magnification, this module ($\mathcal{R}_{\text{SMM}}$) acts as a sparsity rectifier. Operating within the fixed high-resolution space ($Scale=1$), it focuses solely on filtering out upsampling artifacts (e.g., ringing effects) and modulating the energy distribution to fit the sharp Dirac delta prior.

\subsection{\textbf{Isomorphic Source-Generation Unit (ISGU)}}
As shown in Fig.~\ref{fig:FOCUS} (b), the atomic computational engine underpinning both DMM and SMM is the Isomorphic Source-Generation Unit (ISGU). We term it isomorphic because both stages share an identical topological architecture, reflecting our insight that the physical retrieval of a point source, whether from background noise or from calculation artifacts, relies on the same sparsity structure.

Let $\mathbf{Z}_{in}$ be the input feature. The ISGU introduces a structural sparsity induction architecture that bifurcates the flow into:
\begin{equation}
    \mathbf{Z}_{int} = \mathcal{C}_{int}(\mathbf{Z}_{in}), \quad \mathbf{Z}_{mask} = \mathcal{C}_{mask}(\mathbf{Z}_{in}).
\end{equation}
The \textit{Intensity Path} ($\mathbf{Z}_{int}$) estimates the potential radiant energy, while the \textit{Mask Path} ($\mathbf{Z}_{mask}$) estimates the spatial existence probability. The final sparse output is generated via a gated projection:
\begin{equation}
    \mathbf{Z}_{out} = \mathcal{UP}_{\times s}(\mathbf{Z}_{int}) \odot \varsigma(\mathcal{UP}_{\times s}(\mathbf{Z}_{mask}))
\end{equation}
here, $\mathcal{UP}_{\times s}$ denotes the sub-pixel convolution with scale factor $s$ ($s=3$ for DMM, $s=1$ for SMM), $\varsigma$ is the Sigmoid activation, and $\odot$ is the Hadamard product. By using the mask to physically gate the intensity, ISGU explicitly enforces structural sparsity at the architectural level.

\subsubsection{\textbf{Dynamic Physics-Constrained}}
To guide the lightweight network towards physically valid solutions, we introduce a composite loss function that orchestrates a dynamic game among three constraints. This process is not a simple summation of errors, but a competitive equilibrium of forces, as visualized in Fig.~\ref{fig:FOCUS} (c--e).

The optimization is fundamentally anchored by the Reconstruction Loss ($\mathcal{J}_{rec}$), for which we employ Focal Mean Squared Error (MSE). Mathematically, it minimizes the variance between the prediction $\hat{\mathbf{x}}$ and the ground truth $\mathbf{x}$:
\begin{equation}
    \mathcal{J}_{rec} = \frac{1}{N} \sum_{i} \underbrace{|\hat{\mathbf{x}}_i - \mathbf{x}_i|^\gamma}_{\text{Focal Weight}} \cdot (\hat{\mathbf{x}}_i - \mathbf{x}_i)^2.
\end{equation}
This term acts as a spatial anchor. 
By penalizing deviations in intensity distribution, it forces the centroid of the predicted spot to align with the ground truth. 
However, to minimize the spatial penalty of slight misalignments, MSE inherently favors a smooth, Gaussian-like distribution over a sharp Dirac delta, leading to the smoothness bias.

To counteract this smoothing and eliminate background noise, the Structural Sparsity Enforcement term ($\mathcal{J}_{sse}$) imposes an $\ell_1$-regularization:
\begin{equation}
    \mathcal{J}_{sse} = \frac{1}{N} \| \hat{\mathbf{x}} \|_1, \quad \text{with gradient } \nabla \mathcal{J}_{sse} = \text{sign}(\hat{\mathbf{x}}).
\end{equation}
Unlike $\ell_2$ penalties where the gradient vanishes as intensity approaches zero, the $\ell_1$ gradient remains constant ($\pm 1$). This exerts a uniform suppressive force across the entire image plane. It aggressively truncates background noise and artifacts to zero, but inevitably, it also applies the same downward pressure on the target signal, causing amplitude attenuation.

Crucially, to resolve the conflict between suppression and fidelity, we introduce the physics-constrained flux alignment ($\mathcal{J}_{pcfa}$):
\begin{equation}
    \mathcal{J}_{pcfa} = \left| \sum_{i} \hat{\mathbf{x}}_i - \mathcal{E}_{ref} \right|, \quad \text{where } \mathcal{E}_{ref} = \sum_{j} \mathbf{y}^{clean}_{j}.
\end{equation}
This term acts as a global flux anchor. Since $\mathcal{J}_{sse}$ has suppressed the background to zero and $\mathcal{J}_{rec}$ has locked the spatial topology to the target location, the system creates a restorative force driven by the energy deficit ($\mathcal{E}_{ref} - \sum \hat{\mathbf{x}}$). 
Because the background regions are strongly inhibited by the sparsity term, the optimization landscape forces this missing energy to be poured back into the only active region allowed by the spatial anchor--the target centroid. Consequently, this structure sharpens the peak and restores physical intensity without causing energy drift to the background.

The final objective represents the equilibrium of these conflicting yet complementary forces:
\begin{equation}
    \mathcal{J}_{total} = \mathcal{J}_{rec} + \lambda_1 \mathcal{J}_{sse} + \lambda_2 \mathcal{J}_{pcfa}.
\end{equation}
This mechanism ensures that the final output is not only spatially accurate but also physically sparse and energetically conserved.

\section{Experiment} \label{sec:Experiment}

In this section, we conduct a comprehensive evaluation of the proposed \textbf{FOCUS} framework. 
The core objective is to verify whether a lightweight one-stage direct inversion paradigm can transcend the limitations of the traditional unfolding paradigm. 
We argue that FOCUS not only maintains the physical precision of iterative solvers but also achieves a significant leap in inference efficiency.

\subsection{Experimental Settings}

\textbf{Metrics:} 
The evaluation considers both unmixing fidelity and computational efficiency to assess the performance of the proposed one-stage paradigm. 
To quantify the real time capability and architectural ligthweightness, inference throughput in frames per second (FPS) and floating point operations (FLOPs) are reported.
To assess the fidelity of sub-pixel unmixing, we adopt the Mean Average Precision ($mAP$) as the primary evaluation metric, following the protocol in \cite{han2025dista}. 
The $mAP$ is defined as the mean of Average Precision ($AP_d$) calculated at five sub-pixel distance thresholds: $d \in \{0.05, 0.10, 0.15, 0.20, 0.25\}$ pixels. 
This multi-threshold approach strictly penalizes both spatial localization deviations and radiant intensity estimation errors, reflecting the one-to-many decomposition accuracy. 

\textbf{Training Settings:} 
The FOCUS framework is implemented in PyTorch and optimized via the composite loss $\mathcal{J}_{total} = \mathcal{J}_{rec} + \lambda_1 \mathcal{J}_{sse} + \lambda_2 \mathcal{J}_{pcfa}$. 
Through cross-validation, the penalty coefficients are set to $\lambda_1 = 1 \times 10^{-4}$ for structural sparsity and $\lambda_2 = 0.1$ for flux calibration. 
The model is trained for 300 epochs using the Adam optimizer with a batch size of 64 and the initial learning rate is $1 \times 10^{-3}$. 
To ensure the fairness of the throughput comparison, all models (including baselines) are evaluated on a single NVIDIA GeForce RTX 4090 GPU.

\begin{figure}[h]
    \flushright 
    \includegraphics[width=.48\textwidth]{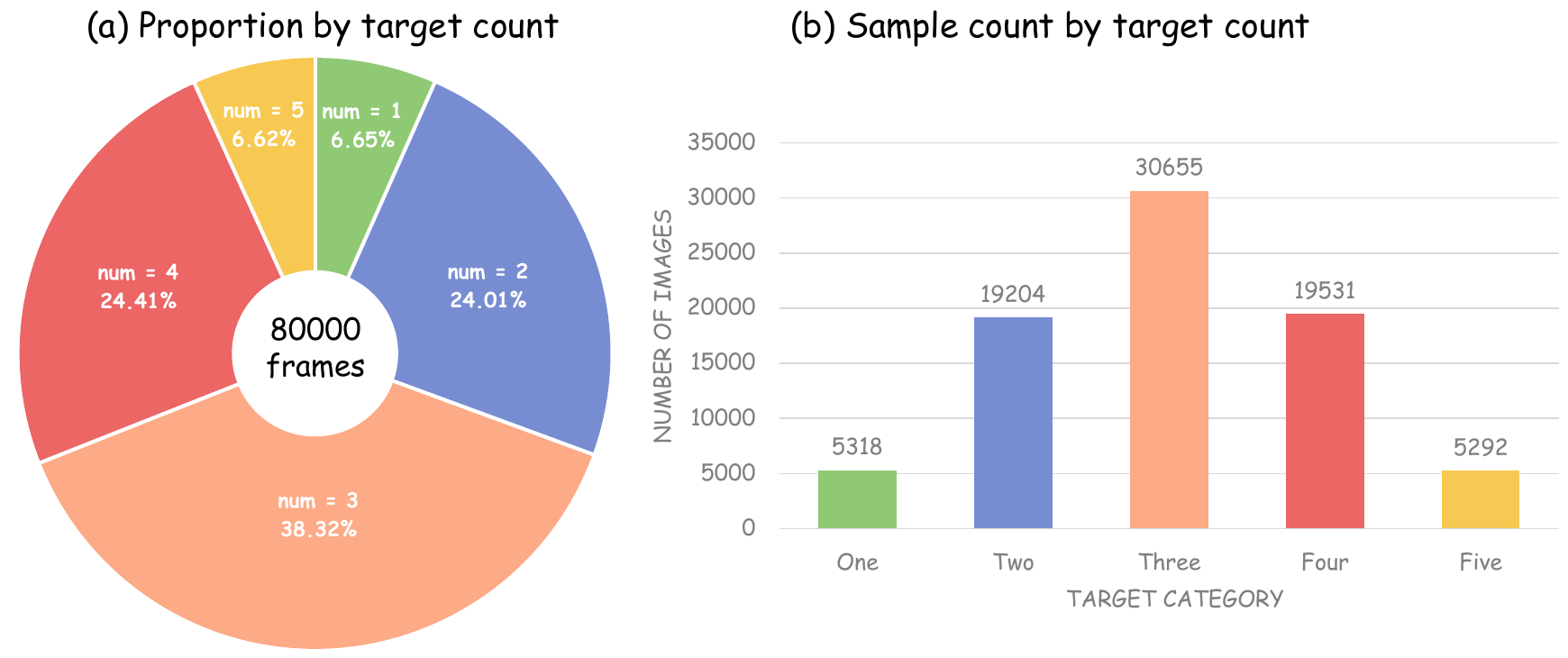}
    \caption{Target count distribution of the CSIST-100K training set. \textbf{(a)} Proportion by target count. \textbf{(b)} Absolute sample count per category. 
    The distribution follows an approximate binomial pattern, with 2--4 target scenarios dominating (86.7\%).}
    \label{fig:CSIST_100K}
\end{figure}

\textbf{Dataset:}
We evaluate on the CSIST-100K dataset \cite{han2025dista}, a large-scale synthetic benchmark for CSIST Unmixing comprising 80,000 training samples, 10,000 validation samples, and 10,000 test samples. 
Each sample consists of an $11 \times 11$ low-resolution infrared observation and a corresponding $33 \times 33$ high-resolution ground truth, synthesized via a Gaussian PSF degradation model with $\sigma = 0.5$. 
Beyond the original data description, we further analyze the target count distribution, finding that it follows an approximate binomial pattern (Fig.~\ref{fig:CSIST_100K}): 
2--4 target scenarios dominate (86.7\%), consistent with the realistic frequency of closely-spaced targets within a sub-pixel distance, while single-target and five-target scenarios are retained at approximately 6.6\% each to enhance generalization across density extremes. 
The validation and test sets follow the same distribution.

\subsection{Comparison with State-of-the-Arts}

\begin{table*}[t]
\centering
\caption{Quantitative comparison with State-of-the-Art Deep Unfolding and Super-Resolution paradigms on the CSIST-100K dataset. 
\textbf{Bold} and {*} values denote the best results and the performance upper bound, respectively. They are highlighted in the bottom row.}
\label{tab:results}
\resizebox{\textwidth}{!}{
\begin{tabular}{l|c|ccccc|c|c}
\hline
\textbf{MODEL} & \textbf{mAP} & \textbf{$AP_{05}$} & \textbf{$AP_{10}$} & \textbf{$AP_{15}$} & \textbf{$AP_{20}$} & \textbf{$AP_{25}$} & \textbf{FPS} & \textbf{FLOPs} \\ \hline
\rowcolor[HTML]{F2F2F2} \multicolumn{9}{c}{\textit{Deep Unfolding Paradigms for CSIST Unmixing}} \\ \hline
LIHT \cite{wang2016learning_LIHT}& 10.35 & 0.06 & 0.92 & 4.99 & 14.74 & 30.50 & 46009 & 1.36 G \\
LAMP \cite{metzler2017learned_LAMP}& 14.22 & 0.05 & 1.11 & 7.31 & 21.56 & 41.06 & 12625 & 0.28 G \\
LISTA \cite{gregor2010learning_LISTA}& 30.13 & 0.25 & 4.13 & 22.29 & 51.18 & 72.82 & 34111 & 1.36 G \\
ISTANet \cite{zhang2018istanet_ISTANet}& 44.99 & 0.40 & 7.10 & 40.10 & 82.60 & 94.80 & 2127 & 11.37 G \\
ISTANet+ \cite{zhang2018istanet_ISTANet}& 45.57 & 0.30 & 6.70 & 40.40 & 84.30 & 96.00 & 1696 & 24.33 G \\
DISTA-Net \cite{han2025dista}& 46.47 & 0.30 & 6.70 & 41.20 & 86.40 & 97.60 & 10992 & 35.10 G \\
FISTANet \cite{xiang2021fista_FISTANet}& 44.97 & 0.50 & 7.70 & 40.40 & 81.80 & 94.50 & 12516 & 18.96 G \\
TiLISTA \cite{liu2019alista_TiLISTA}& 14.95 & 0.06 & 1.23 & 7.72 & 22.50 & 46.23 & 9408 & 0.28 G \\
USRNet \cite{zhang2020deep_USRNet}& 41.82 & 0.30 & 6.80 & 35.70 & 75.70 & 90.50 & 1084 & 36.02 G \\
RPCANet \cite{wu2024rpcanet_RPCANet}& 45.47 & 0.30 & 6.60 & 40.20 & 83.80 & 96.40 & 1892 & 47.39 G \\ \hline
\rowcolor[HTML]{F2F2F2} \multicolumn{9}{c}{\textit{Modified Image Super-Resolution Paradigms for CSIST Unmixing}} \\ \hline
RLFN \cite{kong2022residual}& 29.03 & 0.20 & 3.40 & 20.00 & 48.60 & 73.00 & 22330 & 3.86 G \\
SMFANet \cite{smfanet}& 28.01 & 0.20 & 3.40 & 19.60 & 47.10 & 69.80 & 12889 & 1.29 G \\
CGA \cite{waleed2026cga}& 45.22 & 0.40 & 7.20 & 39.70 & 83.00 & 95.90 & 1784 & 11.95 G \\
RDN \cite{zhang2018residual_RDN}& 45.57 & 0.30 & 7.20 & 40.60 & 83.50 & 96.20 & 5050 & 0.17 T \\
EDSR \cite{lim2017enhanced_EDSR}& 45.41 & 0.40 & 7.50 & 41.00 & 82.90 & 95.30 & 28075 & 3.02 G \\
ASSA \cite{qian2021assanet}& 43.50 & 0.30 & 6.30 & 37.20 & 79.70 & 94.00 & 2138 & 12.74 G \\
PFT \cite{long2025progressive}& 45.34 & 0.40 & 8.10 & 39.50 & 82.30 & 96.40 & 1076 & 9.78 G \\
SSIU \cite{ni2025structural_SSIU}& 43.73 & 0.40 & 7.00 & 38.90 & 79.70 & 92.70 & 7187 & 0.46 G \\ 
CAMixerSR \cite{wang2024_CAMixerSR}& 44.87 & 0.40 & 7.20 & 40.10 & 82.30 & 94.40 & 1138 &  20.42G \\ \hline
\rowcolor[HTML]{E6F2FF} \textbf{FOCUS (Ours)} & 45.50 & 0.40 & 7.30 & 41.00 & 83.20 & 95.50 & \textbf{122522} & 1.63 G \\ 
\rowcolor[HTML]{E6F2FF} \textbf{FOCUS+ (Ours){*}} & \textbf{46.53} & 0.40 & 7.40 & 42.20 & 85.60 & 97.00 & 11003 & 41.49 G \\ \hline
\end{tabular}
}
\end{table*}

\subsubsection{\textbf{Comparison with Deep Unfolding Networks}}
The upper part of Table~\ref{tab:results} evaluates FOCUS against the deep unfolding paradigm. 
\textbf{ISTANet} is utilized as the foundational baseline, as it represents the prototypical recursive architecture that governs current CSIST unmixing research. 
Although recent extensions like DISTANet introduce dynamic parameters to improve accuracy, \textbf{the fundamental computational tax remains an inherent property of the underlying ISTANet based recursion.} 
By comparing FOCUS to this foundational framework, the \textbf{approx. 60$\times$ speedup (122,522 vs 2,127 FPS)} highlights the efficiency leap provided by eliminating the iterative loop itself. 
While standard FOCUS yields a competitive mAP of 45.50\%, the iterative variant FOCUS+ is evaluated to explore the performance ceiling. 
At an equivalent throughput to DISTANet (approx. 11,000 FPS), \textbf{FOCUS+ achieves a higher mAP compared to DISTANet.} 
This verifies that the parameter driven isomorphic logic, specifically the integration of adaptive thresholds and learnable step sizes, provides a more precise representation for source recovery than traditional unrolling units. 
The results confirm that unfolding is not essential for achieving high unmixing fidelity, provided the network logic is guided by appropriate structural constraints. 
This efficiency gain establishes FOCUS as a viable solution for high speed infrared sensing where traditional iterative loops become the primary delay.

\subsubsection{\textbf{Comparison with Modified Super-Resolution Paradigms}} 
Compared to modified image SR paradigms, where the label spaces, loss functions, and evaluation metrics have been reformulated for the unmixing task, FOCUS establishes a superior balance among precision, computational cost, and inference throughput.
Standard reconstruction models are inherently optimized for visual textures, which leads to a smoothness bias that merges discrete radiant sources into continuous energy clusters. 
While modern architectures such as CAMixerSR and EDSR achieve resolution enhancement, they often lag in unmixing fidelity due to the lack of explicit sparsity constraints. 
For example, CAMixerSR yields an mAP of 44.87\%, which is lower than FOCUS despite requiring a much higher computational budget (20.42G vs 1.63G FLOPs). 
Crucially, the throughput of FOCUS reaches 122,522 FPS, \textbf{which is over 100$\times$ faster than CAMixerSR (1,138 FPS) and 60$\times$ faster than the precise CGA scheme (1,784 FPS)}. 
Even when compared to lightweight models like SSIU, \textbf{FOCUS maintains a 17$\times$ speed advantage while providing a nearly 2\% gain in mAP.} 
On the other end of the spectrum, models like RLFN fail to capture the sparse nature of infrared targets despite their high speed, resulting in a sever drop in mAP compared to FOCUS. 
The superiority of FOCUS is attributed to the ISGU architecture, where the mask path explicitly gates the intensity path to enforce structural sparsity. 
This design represents a successful adaptation of reconstruction principles for the inversion of discrete point sources. 
By injecting task specific constraints into a feedforward flow, FOCUS allows a 1.63G FLOPs model to resolve aliased energy regions with accuracy and speed that outperform both heavy iterative and visual restoration paradigms, proving that architectural efficiency does not necessitate a sacrifice in source recovery precision.


\begin{figure*}[htbp]
    \centering
    \includegraphics[width=.98\textwidth]{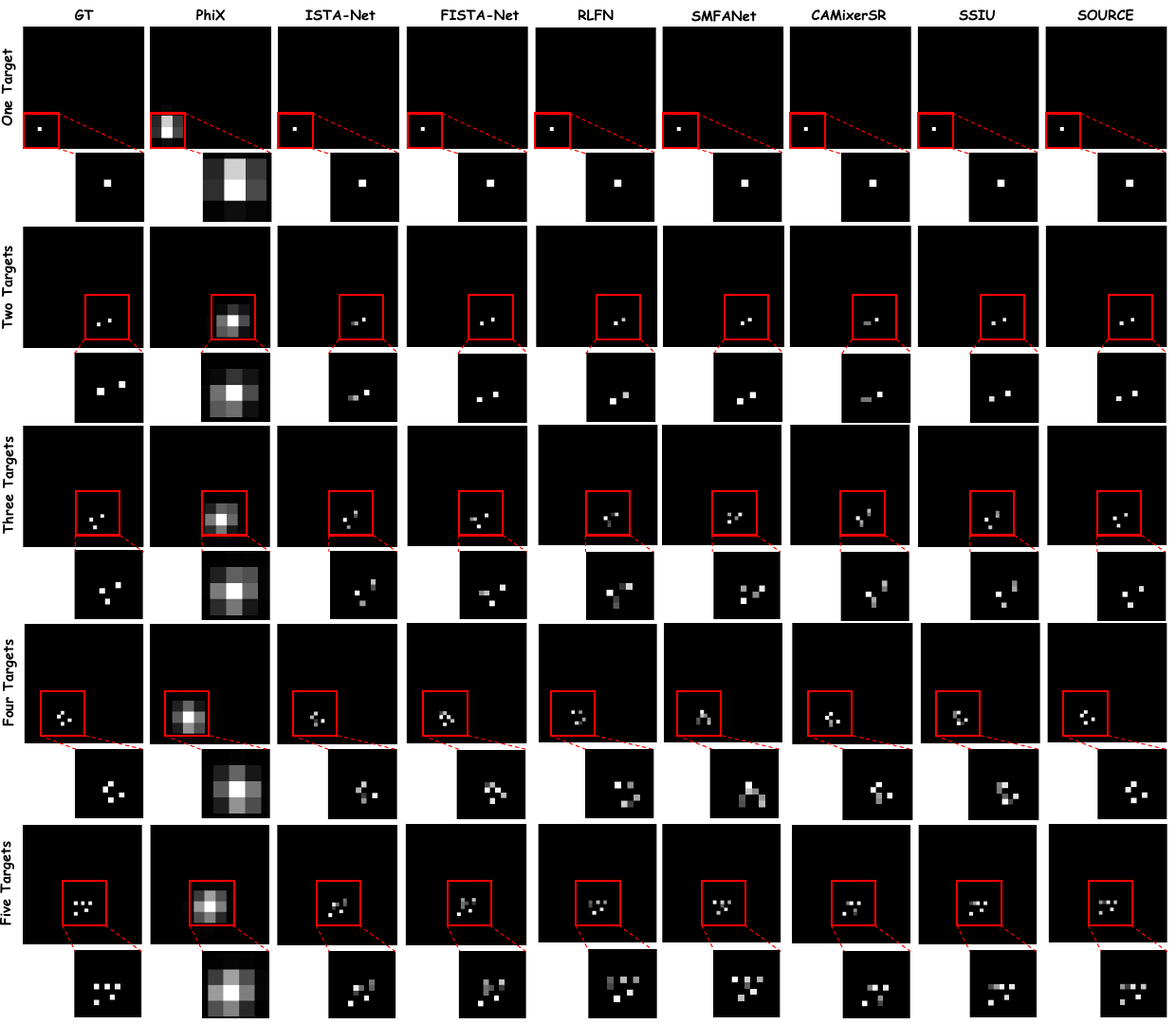}
    \caption{Visual comparison of unmixing results on the CSIST-100K dataset across varying target densities. Each row represents a specific scenario ranging from a single target to five congested targets, while the columns display the reconstruction outcomes of different paradigms. The red boxes highlight the zoomed-in details of sub-pixel target clusters. These results confirm that the proposed framework maintain robust unmixing performance and precise localization even in highly congested scenarios.}
    \label{fig:visual_results}
\end{figure*}

\subsection{Visual Analysis}
Fig.~\ref{fig:visual_results} presents the qualitative comparison of reconstruction results on the CSIST-100K dataset across different target densities. 
Overall, FOCUS achieves a significant reduction in model parameters while maintaining high reconstruction accuracy, a balance that is visually apparent across all test scenarios.

Specifically, the single-target results indicate that the resolution enhancement provided by FOCUS does not introduce spatial distortions or compromise the localization accuracy of isolated infrared targets. 
This demonstrates that the model effectively preserves the point-like characteristics of infrared signals during the upscaling process.

In comparison with deep unfolding methods such as ISTA-Net and FISTA-Net, FOCUS exhibits higher precision in recovering target intensities and sharp edges. 
Furthermore, a distinct performance gap emerges when comparing FOCUS to SR methods like ESPCN, SRCNN, and CAMixerSR. 
As the number of aliased targets increases from three to five, these SR-based approaches suffer from severe patchy artifacts and blurred energy distributions, failing to distinguish closely spaced targets. 
In contrast, FOCUS consistently produces clear and distinct reconstructions even in highly congested scenarios.
In summary, these visual results confirm that FOCUS effectively balances model efficiency with high-fidelity reconstruction. 
It remains robust against aliasing interference, making it a reliable solution for infrared compressed sensing reconstruction.


\subsection{Ablation Study}

We structure the ablation study along three progressive dimensions: the rationality of architectural design, the effectiveness of physics-guided constraints, and the generalization capability across diverse scenarios.

\begin{table}[htbp]
\renewcommand\arraystretch{1.6}
\footnotesize
\centering
\centering
\caption{Ablation study of the proposed components on the CSIST-100K dataset.}
\label{tab:ablation}
\begin{tabular}{cccc|ccc}
\hline
\textbf{Baseline} & \textbf{SSE} & \textbf{PGFC} & \textbf{HCF} & \textbf{$mAP$ (\%)} & \textbf{$AP_{20}$} & \textbf{$AP_{25}$}\\ \hline
$\checkmark$    &    &    &   & 42.61   & 77.30 & 89.70          \\
$\checkmark$    & $\checkmark$ &  &  & 43.74 (+1.13)    & 79.30 &  90.00     \\
$\checkmark$    & $\checkmark$ & $\checkmark$  &  & 44.10 (+0.36) & 79.90 & 91.30 \\
$\checkmark$    & $\checkmark$ & $\checkmark$  & $\checkmark$ & \textbf{45.50 (+1.40)}  & \textbf{83.20} & \textbf{95.60}\\ \hline
\end{tabular}
\end{table}

\subsubsection{\textbf{Impact of Different Designs}} 
To evaluate the individual contribution of each component within FOCUS, an incremental ablation study is conducted on the CSIST 100K dataset. 
A generic one-stage regression network serves as the baseline, yielding an mAP of 42.61\% and an $AP_{25}$ of 89.70\%. 
The results summarized in Table~\ref{tab:ablation} verify that each design element is necessary for resolving subpixel radiant sources.

The integration of Structural Sparsity Enforcement (SSE) through the ISGU mask path yields a 1.13\% mAP increase. 
This mechanism introduces a suppressive force that pushes low intensity background clutter toward zero, facilitating the formation of sharper target centroids. 
The benefit of this sparsity induction is reflected in the sub-pixel localization precision, where the $AP_{20}$ score rises from 77.30\% to 79.30\%. 
By penalizing the absolute sum of output intensities, SSE ensures that the network prioritizes a sparse solution to counter the energy diffusion typical of convolutional layers. 
However, the downward gradient of the L1 norm may also leads to target signal attenuation alongside background suppression.

Physics Guided Flux Calibration (PGFC) resolves this attenuation by anchoring radiant energy to the reference level. 
By forming an equilibrium with the suppressive force of SSE, PGFC reallocates energy back to the identified target coordinates. 
This process improves mAP to 44.10\% and increases $AP_{25}$ to 91.30\%. 
This outcome confirms that flux conservation is necessary to maintain the radiant fidelity of discrete sources while enforcing sparsity. 
The interaction between SSE and PGFC ensures that the model sharpens the target response without sacrificing its radiant magnitude.

\begin{table}[htbp]
\renewcommand\arraystretch{1.4}
\footnotesize
\centering
\caption{Average intensity deviation between TP detections and GT under different constraint configurations. Lower absolute value indicates better amplitude recovery.}
\label{tab:ablation_intensity}
\begin{tabular}{lcccc}
\toprule
\multirow{2}{*}{\textbf{Constraints}} & \multicolumn{2}{c}{\textbf{$\Delta$Amplitude}} & \multicolumn{2}{c}{\textbf{Relative Improvement}} \\
\cmidrule(lr){2-3} \cmidrule(lr){4-5}
& $AP_{05}$ & $AP_{10}$ & $AP_{05}$ & $AP_{10}$ \\
\midrule
$\mathcal{L}_{MSE}$  & -66.223 & -68.061 & --      & --     \\
$\mathcal{L}_{Sparsity}$  & -52.478 & -54.873 & 20.76\% ($\uparrow$) & 19.38\% ($\uparrow$)\\
$\mathcal{L}_{Energy}$  & \textbf{-49.036} & \textbf{-53.366} & 6.56\%  ($\uparrow$)& 2.75\% ($\uparrow$) \\
\bottomrule
\end{tabular}
\end{table}
The introduction of the HCF flow yields the most measurable performance gain, raising the mAP by 1.40\% to reach 45.50\%. 
This strategy decouples coarse spatial topology recovery from subpixel coordinate refinement into dedicated modules. 
Isolating geometric estimation from artifact suppression reduces optimization entanglement compared to undifferentiated one-stage architectures. 
The efficacy of this stratified approach is evidenced by the increase in strict localization metrics, where $AP_{20}$ reaches 83.20\% and $AP_{25}$ attains 95.60\%. 
This hierarchy ensures that the network logic converges toward a sharp Dirac delta prior. 
Collectively, these results justify the transition from iterative unfolding to a task specific direct mapping paradigm.

\subsubsection{\textbf{Impact of Constraints}}
Table~\ref{tab:ablation_intensity} examines the amplitude recovery fidelity of detections relative to the ground truth under progressive constraint integration. 
\begin{figure}[htbp]
    \centering
    \includegraphics[width=.48\textwidth]{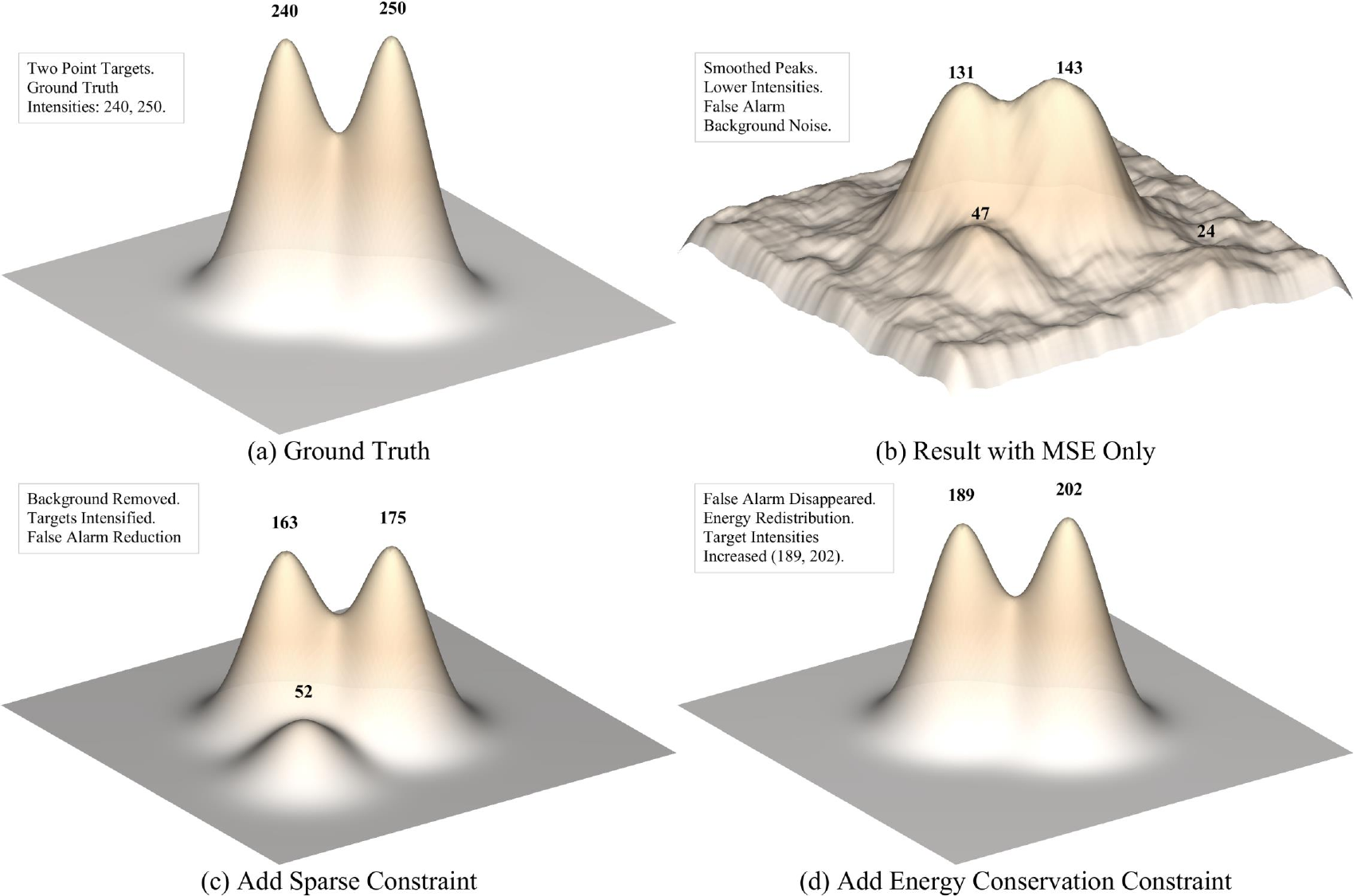}
    \caption{Changes in energy intensity after gradually adding constraints.}
    \label{fig:constraints}
\end{figure}
The individual impact of these constraints is visualized through a representative case study in Fig.~\ref{fig:constraints}. 
Theoretically, sparsity regularization exerts a uniform downward pressure that would attenuate signal intensities if utilized in isolation. 
However, as an auxiliary constraint alongside the spatial anchoring of mean square error, sparsity enforcement primarily suppresses diffuse background noise and low intensity false alarms. 
This competitive interaction forces the network to concentrate radiant flux toward the predicted target coordinates, which results in the intensity growth observed in Table~\ref{tab:ablation_intensity}. 
Specifically, the amplitude error at the 0.05 pixel threshold improves from 66.22\% to 52.48\%, representing a 20.76\% gain.

Subsequently, flux conservation provides targeted energy calibration to further eliminate residual false alarms. 
By anchoring the total predicted flux to the reference level, this constraint facilitates the reallocation of energy from suppressed artifacts back to the target centroids. 
This restorative process reduces the amplitude deficit by an additional 6.56\% at the 0.05 pixel threshold and 2.75\% at the 0.10 pixel threshold. 
The diminishing returns at relaxed thresholds are consistent with the focus of flux conservation on fine grained amplitude fidelity rather than coarse localization. 
While the 3D visualizations in Fig.~\ref{fig:constraints} illustrate this energy redistribution for a specific instance, the statistical data in Table~\ref{tab:ablation_intensity} confirms that these constraints act as complementary forces across the entire dataset. 
The synergy between sparsity and energy conservation establishes a restorative equilibrium that ensures unmixing fidelity by maintaining both spatial discreteness and radiant accuracy.

\begin{table}[htbp]
\renewcommand\arraystretch{1.4}
\footnotesize
\centering
\caption{Robustness evaluation under different Gaussian noise levels.}
\label{tab:gaussian_noise}
\begin{tabular}{c|cccccc}
\hline
$\sigma$ & FOCUS & ASSA & CAMixer & CGA & ISTANet & USRNet \\ \hline
0    & \textbf{45.50} & 43.50 & 44.87 & 45.22 & 44.99 & 41.82 \\
0.05 & \textbf{45.45} & 36.44 & 44.80 & 20.93 & 44.94 & 41.74 \\
0.1  & \textbf{45.37} & 36.43 & 44.61 & 19.92 & 44.81 & 41.51 \\
0.3  & \textbf{44.51} & 35.88 & 42.98 & 17.58 & 43.72 & 39.24 \\
0.6  & \textbf{41.87} & 34.54 & 38.48 & 16.12 & 40.24 & 32.48 \\
1.0  & \textbf{37.12} & 31.53 & 33.10 & 14.98 & 34.89 & 23.04 \\ \hline
\end{tabular}
\end{table}

\begin{figure}[htbp]
    \centering
    \includegraphics[width=\linewidth]{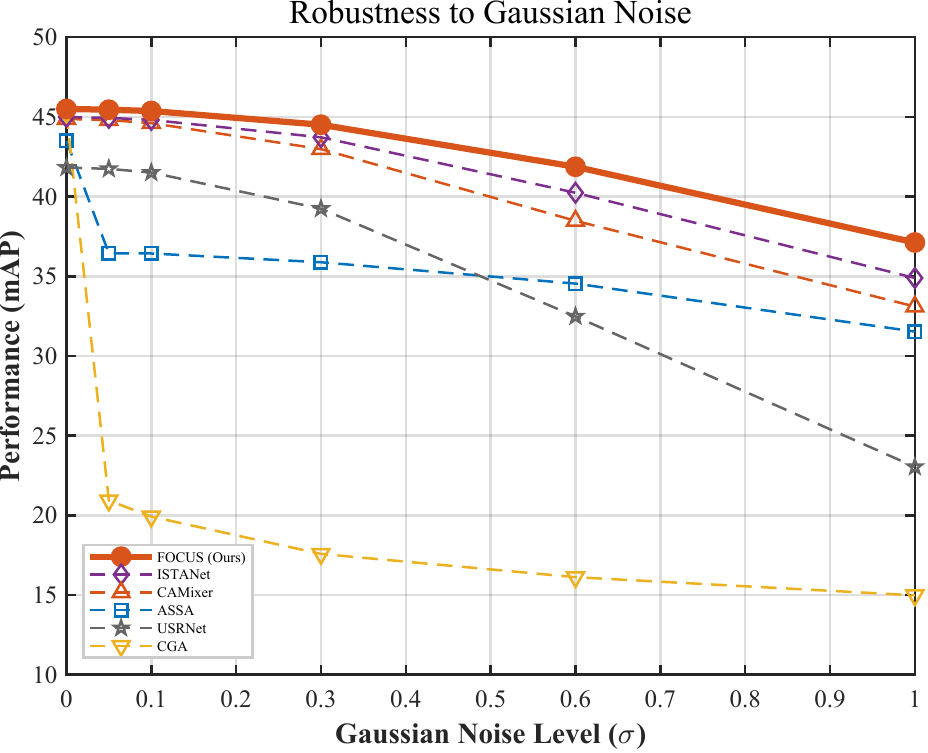}
    \caption{Robustness to Gaussian noise. FOCUS maintains a consistently higher performance trajectory across all noise levels, while competing methods exhibit steeper degradation beyond the training distribution.}
    \label{fig:gaussian_noise}
\end{figure}

\subsubsection{\textbf{Generalization--Robustness to Gaussian Noise}}
The resilience of unmixing schemes against additive noise corruption is evaluated by injecting Gaussian noise into test inputs with standard deviations $\sigma$ ranging from 0 to 1.0. 
This selection spans from ideal observations to degraded conditions that extend far outside the training distribution, which is limited to $\sigma \leq 0.05$. 
The quantitative results summarized in Table~\ref{tab:gaussian_noise} and the performance trajectories illustrated in Fig.~\ref{fig:gaussian_noise} confirm that FOCUS maintains a consistent lead across all noise intensities. 
Under noise free conditions ($\sigma=0$), FOCUS achieves an mAP of 45.50\%, which is superior to both the best reconstruction model CAMixer and the strongest unfolding baseline ISTANet.

A detailed analysis of the degradation curves reveals a clear divergence in generalization capability between the proposed paradigm and existing models. 
As noise intensity increases beyond the training boundary, all competing methods exhibit steep performance drops. 
The fragility of generic architectures is most evident in CGA, which suffers an immediate precision collapse from 45.22\% at $\sigma=0$ to 20.93\% at $\sigma=0.05$, eventually reaching a floor of 14.98\%. 
Similarly, SR based models such as USRNet and ASSA show rapid downward trajectories, indicating that pure data driven fitting fails to maintain signal integrity under out of distribution perturbations. 
In contrast, FOCUS demonstrates a characteristic of graceful degradation, preserving an mAP of 37.12\% even at the highest noise level. 

This robust performance indicates that the integration of radiant flux conservation and structural sparsity provides a restorative equilibrium that prevents the unmixing logic from failing. 
By anchoring the optimization to radiant ground truth levels rather than relying solely on learned textural patterns, FOCUS effectively distinguishes radiant target spikes from stochastic background fluctuations. 
The widening performance gap between FOCUS and the next best competitor, ISTANet, at $\sigma=1.0$ (37.12\% vs 34.89\%) further confirms the advantage of the onestage mapping architecture in ensuring consistent unmixing fidelity. 
Collectively, these results verify that a well constrained direct mapping paradigm can provide stable source recovery across diverse sensor environments without the requirement for iterative solvers or retraining.

\begin{table}[htbp]
\renewcommand\arraystretch{1.4}
\footnotesize
\centering
\caption{Robustness evaluation under different Poisson noise levels.}
\label{tab:poisson_noise}
\begin{tabular}{c|cccccc}
\hline
Peak & FOCUS & ASSA & CAMixer & CGA & ISTANet & USRNet \\ \hline
10000 & \textbf{44.38} & 42.28 & 43.54 & 43.39 & 43.85 & 40.42 \\
5000  & \textbf{43.36} & 41.06 & 42.35 & 41.93 & 42.73 & 39.11 \\
1000  & \textbf{35.98} & 34.03 & 34.98 & 32.98 & 35.71 & 31.65 \\
500   & \textbf{30.08} & 28.41 & 29.22 & 27.08 & 30.01 & 25.89 \\
255   & \textbf{23.58} & 21.79 & 23.17 & 21.49 & 23.02 & 20.09 \\ \hline
\end{tabular}
\end{table}

\subsubsection{\textbf{Generalization--Robustness to Poisson Noise}}
\begin{figure}[htbp]
    \centering
    \includegraphics[width=\linewidth]{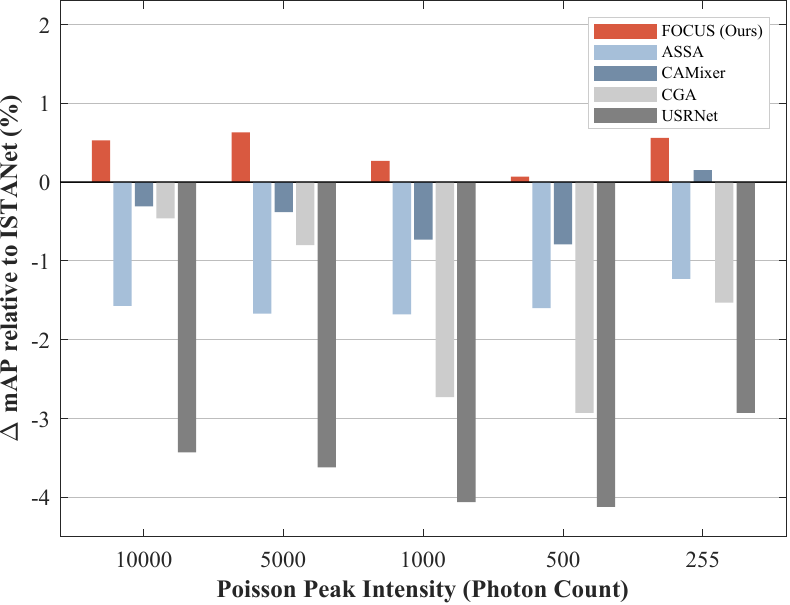}
    \caption{Generalization to Poisson noise. FOCUS consistently leads all competing methods across the full range of peak photon counts, from scientific-grade sensors (peak=10000) to severely photon-starved conditions (peak=255).}
    \label{fig:poisson_noise}
\end{figure}
Poisson noise, which originates from the discrete quantum nature of photon detection, represents the primary source of interference in low light infrared imaging. 
This evaluation considers representative comparison schemes across a spectrum of peak photon counts, ranging from scientific grade sensors at peak 10000 to severely photon starved conditions at peak 255. 
As detailed in Table~\ref{tab:poisson_noise}, FOCUS maintains the highest mAP scores across all noise intensities. 
To further quantify this robustness, Fig.~\ref{fig:poisson_noise} illustrates the performance deviation of various models relative to the ISTANet baseline. 
While ISTANet provides a stable iterative reference, FOCUS consistently produces a positive margin above this baseline, which confirms that the one-stage mapping architecture can generalize reliably across diverse sensor states.

The advantage of the proposed framework becomes increasingly measurable as the photon count decreases. 
At a peak intensity of 1000, FOCUS achieves an mAP of 35.98\%, maintaining a consistent lead over both CAMixer and CGA. 
Under the extreme conditions of peak 255, where the signal to noise ratio is severely compromised, generic reconstruction models such as USRNet and ASSA suffer from precision collapse and fall well below the ISTANet baseline. 
In contrast, FOCUS preserves an mAP of 23.58\%, which represents a 0.56\% gain over the second best iterative model. 
This resilience is attributed to the competitive equilibrium between flux conservation and structural sparsity. By anchoring the total energy to the radiant ground truth, FOCUS prevents the excessive attenuation of target spikes that typically occurs in generic restoration models during background suppression. 
These results collectively verify that the direct mapping paradigm ensures superior signal integrity without the recursive delays of iterative solvers, establishing FOCUS as a robust solution for unmixing under adverse sensing environments.

\begin{table}[htbp]
\renewcommand\arraystretch{1.4}
\footnotesize
\centering
\caption{Unmixing accuracy of FOCUS across different target counts.}
\label{tab:target_density}
\begin{tabular}{c|c|ccccc}
\hline
\textbf{Nums} & $mAP$ & $AP_{05}$ & $AP_{10}$ & $AP_{15}$ & $AP_{20}$ & $AP_{25}$ \\ \hline
1 & 51.10 & 0.60 & 10.40 & 51.50 & 93.20 & 100.00 \\
2 & 49.40 & 0.50 & 9.70  & 47.00 & 90.50 & 99.10  \\
3 & 47.60 & 0.40 & 8.70  & 44.90 & 86.40 & 97.60  \\
4 & 42.70 & 0.30 & 5.90  & 36.60 & 78.00 & 92.60  \\
5 & 38.10 & 0.10 & 4.40  & 31.60 & 68.40 & 86.20  \\ \hline
\end{tabular}
\end{table}
\subsubsection{\textbf{Generalization--Performance Across Target Densities}}

\begin{figure}[htbp]
    \centering
    \includegraphics[width=0.48\textwidth]{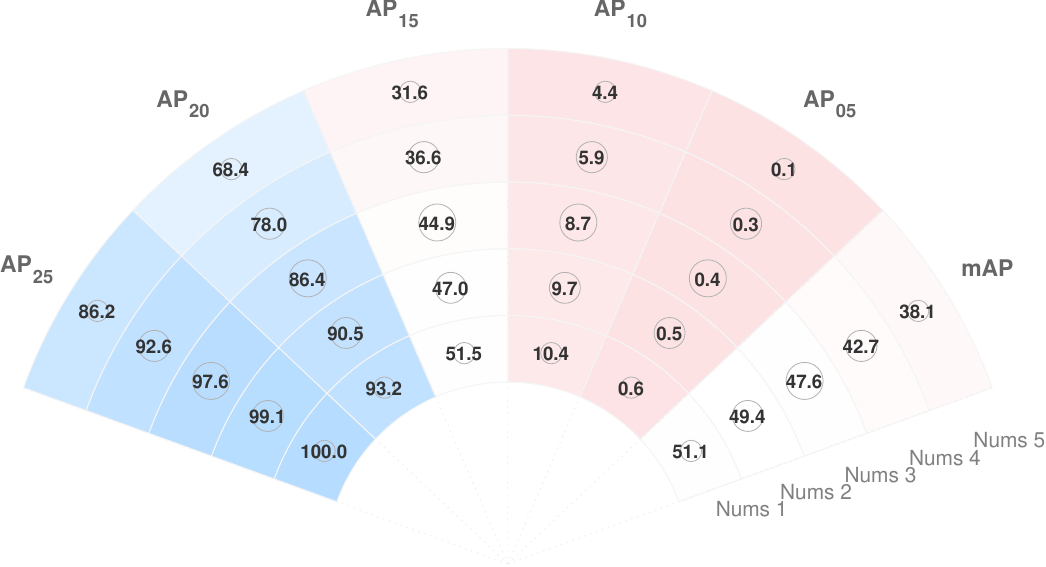}
    \caption{Visualization of unmixing precision across different target counts. The sector map illustrates the distribution of AP scores across five distance thresholds (inner to outer rings) for Nums 1 to 5 (angular segments).}
    \label{fig:target_density_map}
\end{figure}

To evaluate the robustness of FOCUS under varying target densities, we conduct experiments on scenarios containing one to five spatially proximate targets. The quantitative results are detailed in Table~\ref{tab:target_density} and further visualized through the sector heatmap in Fig.~\ref{fig:target_density_map}.

The results indicate that FOCUS yields superior unmixing fidelity across all density levels. 
For isolated targets (Nums 1), the model achieves an mAP of 51.10\% and reaches a perfect 100\% accuracy at the $AP_{25}$ threshold. 
As the target count increases from 1 to 5, the performance exhibits a graceful degradation rather than a sharp collapse. 
Specifically, even in congested scenarios with 5 targets, FOCUS maintains an mAP of 38.10\% and an $AP_{25}$ of 86.20\%. 
This downward trend aligns with the physical nature of the CSIST task, where higher target density leads to severe energy aliasing and overlapping point spread functions, which elevates the difficulty of source separation.

The superiority of the proposed framework is particularly evident in high precision localization. 
As shown in the sector heatmap, while the $AP_{05}$ remains low due to the extreme challenge of 0.05 pixel accuracy, the scores for $AP_{20}$ and $AP_{25}$ remain consistently high even as targets become denser. 
This performance indicates that the isomorphic unit and physics guided constraints effectively concentrate radiant energy into discrete coordinates, preventing the smoothing bias common in standard architectures. 
The ability to resolve 86.20\% of targets within a 0.25 pixel radius under the highest density (Nums 5) confirms that SOURCE provides a reliable solution for congested infrared source recovery.

\begin{table}[htbp]
\renewcommand\arraystretch{1.4}
\footnotesize
\centering
\caption{Performance comparison at higher upsampling factors.}
\label{tab:scale}
\begin{tabular}{l|cc|ccc}
\hline
\textbf{Model} & \textbf{\#P} & \textbf{FLOPs} & \textbf{$mAP$} & \textbf{$AP_{10}$} & \textbf{$AP_{15}$} \\ \hline
\multicolumn{6}{c}{$c=5$} \\ \hline
ISTANet      & 0.171M & 39.54G & 62.57 & 53.8 & 80.5 \\
ISTANet+     & 0.225M & 48.16G & 63.45 & 53.2 & 81.9 \\
FISTANet     & 0.038M & 55.99G & 61.53 & 50.7 & 79.0 \\
FOCUS (Ours) & 0.055M & 4.26G  & \textbf{64.61} & \textbf{54.6} & \textbf{83.7} \\ \hline
\multicolumn{6}{c}{$c=7$} \\ \hline
ISTANet      & 0.171M & 89.51G  & 68.87 & 74.6 & 81.1 \\
ISTANet+     & 0.255M & 103.00G & 71.09 & 74.9 & 84.9 \\
FISTANet     & 0.038M & 0.12T  & 67.81 & 71.7 & 80.3 \\
FOCUS (Ours) & 0.069M & 8.19G  & \textbf{73.28} & \textbf{79.6} & \textbf{86.7} \\ \hline
\end{tabular}
\end{table}

\subsubsection{\textbf{Scalability to Higher Upsampling Factors}}

The resilience of FOCUS is further examined at higher upsampling factors of $c=5$ and $c=7$. 
These scales increase the ill posed nature of the unmixing task by quadratically expanding the latent space required for subpixel localization. 
As reported in Table~\ref{tab:scale}, the onestage paradigm consistently provides superior precision at both resolution settings. 
FOCUS achieves an mAP of 64.61\% at $c=5$ and 73.28\% at $c=7$, which represents a measurable improvement over the strongest unfolding baseline ISTANet+.

A detailed analysis indicates that the precision lead of the proposed framework becomes more distinct as the upsampling factor grows. 
The performance margin between FOCUS and ISTANet+ increases from 1.16\% at $c=5$ to 2.19\% at $c=7$. 
This trend suggests that the coarse to fine hierarchy is more robust to the dimensionality challenges inherent in high resolution mapping compared to the repetitive loops of unfolding architectures. 
By isolating the spatial topology recovery from the coordinate refinement, FOCUS avoids the optimization entanglement that often affects recursive models in highly underdetermined scenarios. 

Furthermore, the computational scaling of FOCUS exhibits a clear advantage over iterative schemes. 
While the FLOPs of baselines such as FISTANet and ISTANet+ escalate sharply, with FISTANet reaching 0.12T at $c=7$, FOCUS maintains a highly efficient footprint of only 8.19G. 
This disparity confirms that the direct mapping paradigm resolves the computational tax of iterative solvers, which typically scale poorly as resolution demands increase. 
The ability of a lightweight network to outperform heavy iterative models while utilizing a fraction of the computational budget verifies the efficacy of the physics guided mapping logic. 
Collectively, these results confirm that FOCUS generalizes gracefully across diverse upsampling factors, maintaining a restorative equilibrium between unmixing fidelity and inference efficiency.

\section{Conclusion} \label{sec:Conclusion}

To resolve the computational overhead and latency characteristic of iterative unfolding methods, this paper introduces \textbf{FOCUS}, a one-stage feed-forward CSIST Unmixing paradigm. 
By identifying the mathematical isomorphism between the degradation models of image SR and CSIST Unmixing, we establish a theoretical basis for adapting efficient convolutional architectures to unmixing. 
Integrating structural sparsity and flux conservation redirects the optimization from visual smoothness toward the precise inversion of discrete radiant sources.
Quantitative evaluations confirm that FOCUS yields an approximate 60$\times$ speedup over the ISTANet baseline, while the unrolled variant FOCUS+ reaches a higher precision record by outperforming current state of the art unfolding models.
This transition from iterative solvers to direct mapping establishes a practical pathway for high speed CSIST Unmixing in time sensitive scenarios.

\bibliographystyle{IEEEtran}
\bibliography{./reference.bib}


\begin{IEEEbiography}[{\includegraphics[width=1.3in,height=1.25in,clip,keepaspectratio]{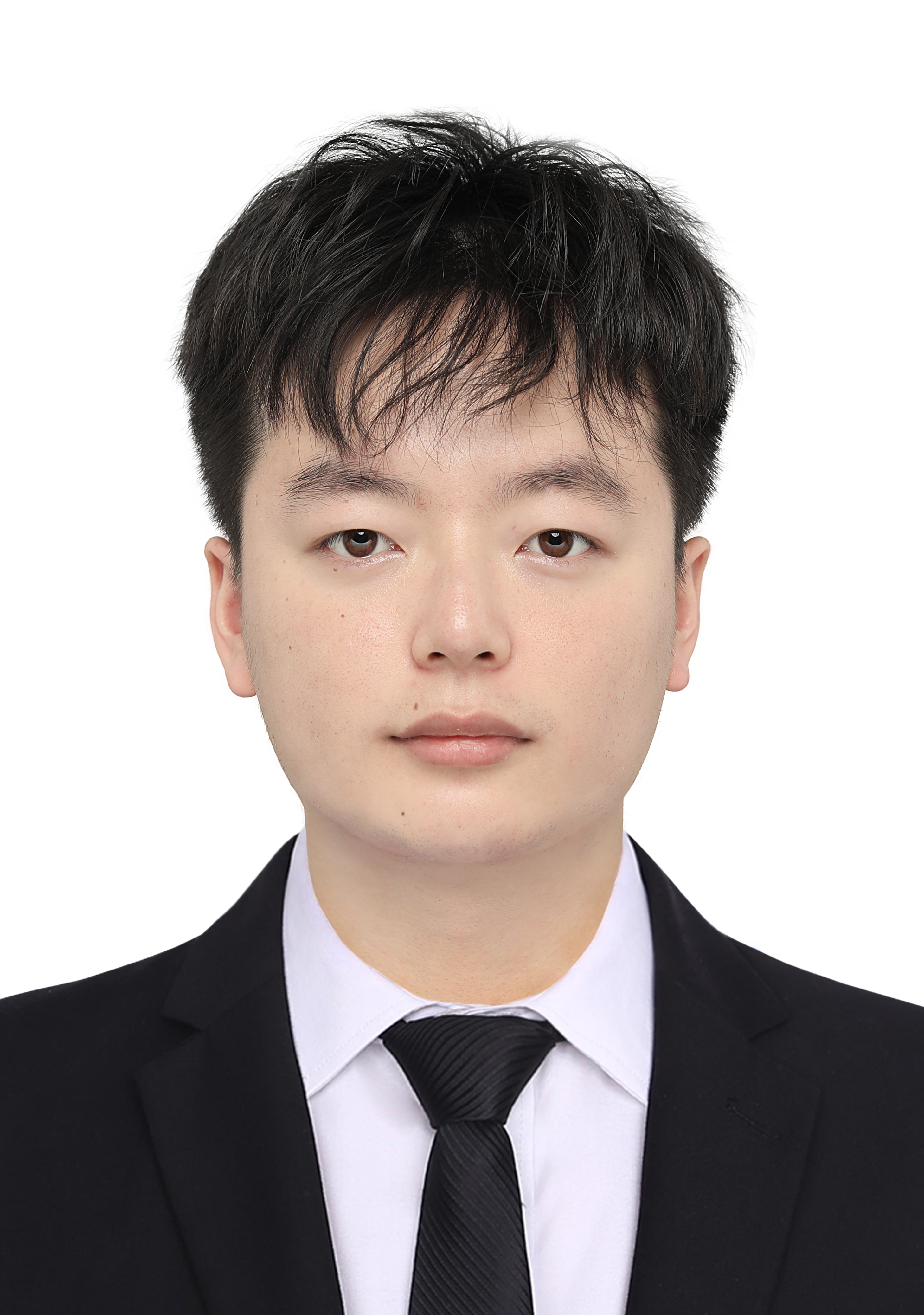}}]{Ximeng Zhai} received the B.S. degree from Tianjin University. He is currently a Ph.D. Candidate at the Xi'an Institute of Optics and Precision Mechanics (XIOPM), University of Chinese Academy of Sciences, China. 

His research interests encompass infrared image processing and object detection.
\end{IEEEbiography}

\begin{IEEEbiography}[{\includegraphics[width=1.3in,height=1.25in,clip,keepaspectratio]{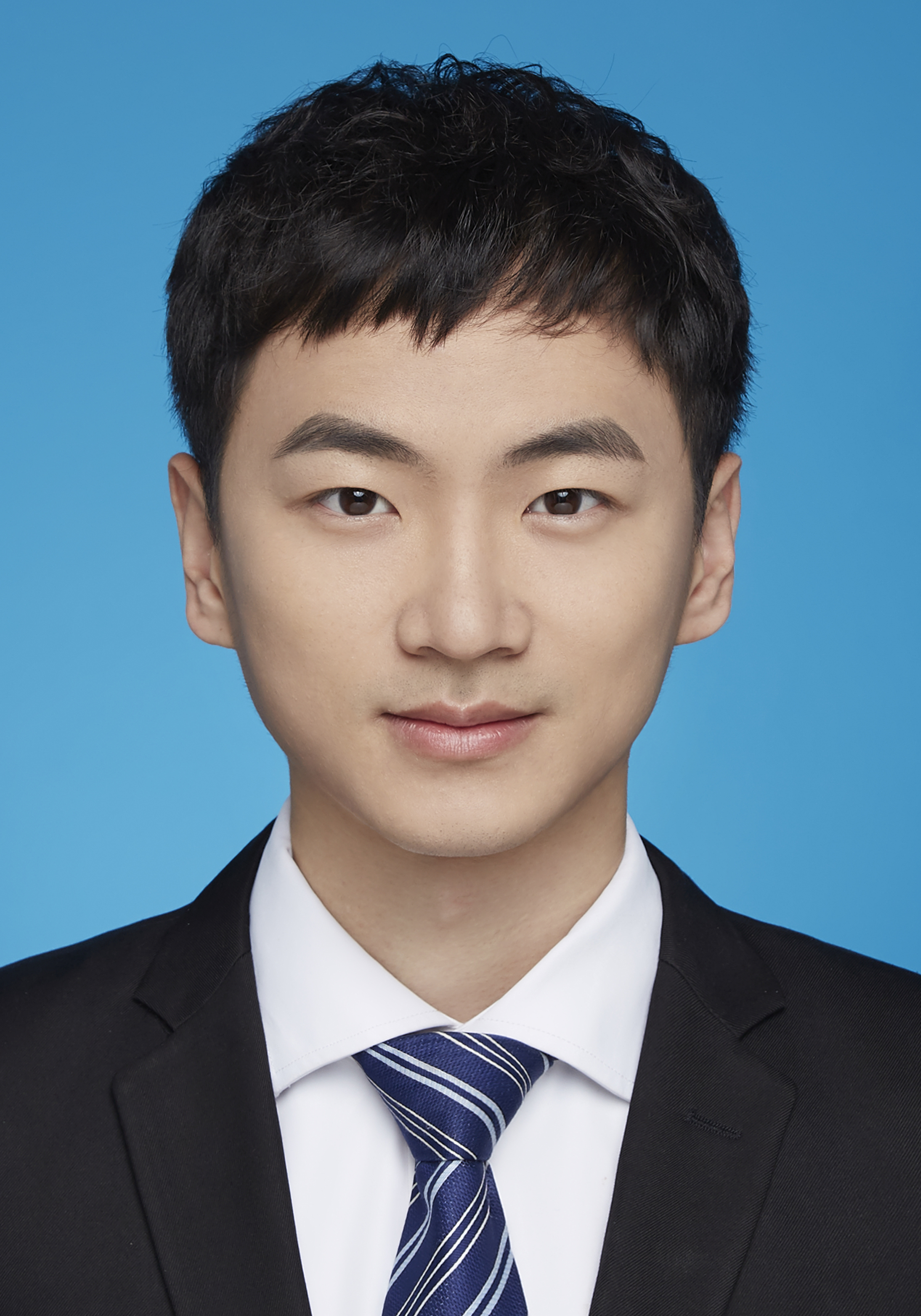}}]{Zheng Wang} received the B.S. degree from Beijing Institute of Technology. 

He currently works at the Beijing Institute of Astronautical Systems Engineering.
His research interests include measurement system and TT\&C systems design.
\end{IEEEbiography}



\begin{IEEEbiography}[{\includegraphics[width=1in,height=1.35in,clip,keepaspectratio]{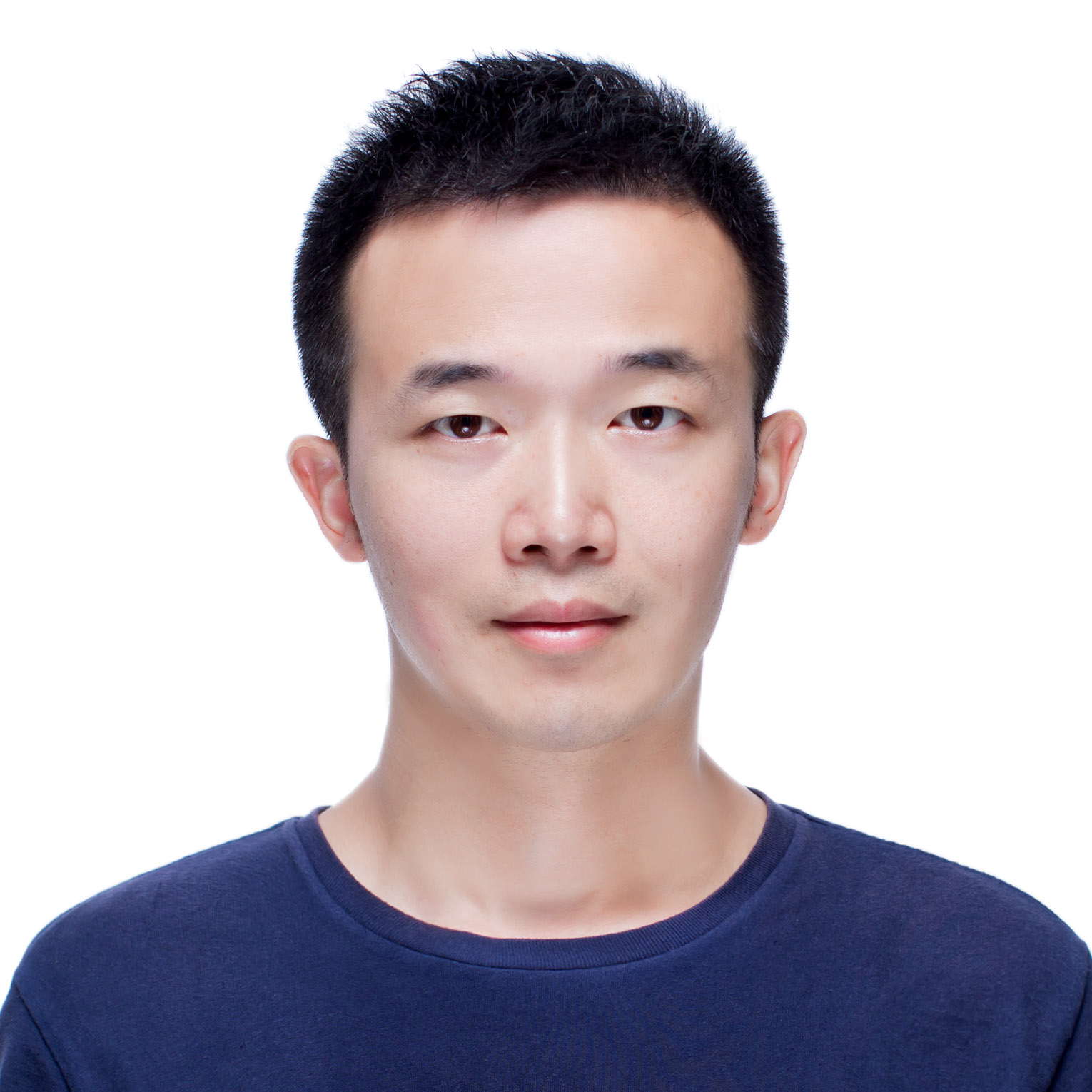}}]{Yaohong Chen} received his M.S. and Ph.D. degree from Xi'an Jiaotong University and University of Chinese Academy of Sciences in 2013 and 2022, respectively. 

He was a visiting scholar in the Department of Electrical and Computer Engineering of Johns Hopkins University during the 2019 to 2020. He is currently an associate research fellow with the Xi'an Institute of Optics and Precision Mechanics, Chinese Academy of Sciences. His research interests include the infrared imaging systems, infrared image processing and infrared imaging circuit.
\end{IEEEbiography}

\begin{IEEEbiography}[{\includegraphics[width=1.3in,height=1.35in,clip,keepaspectratio]{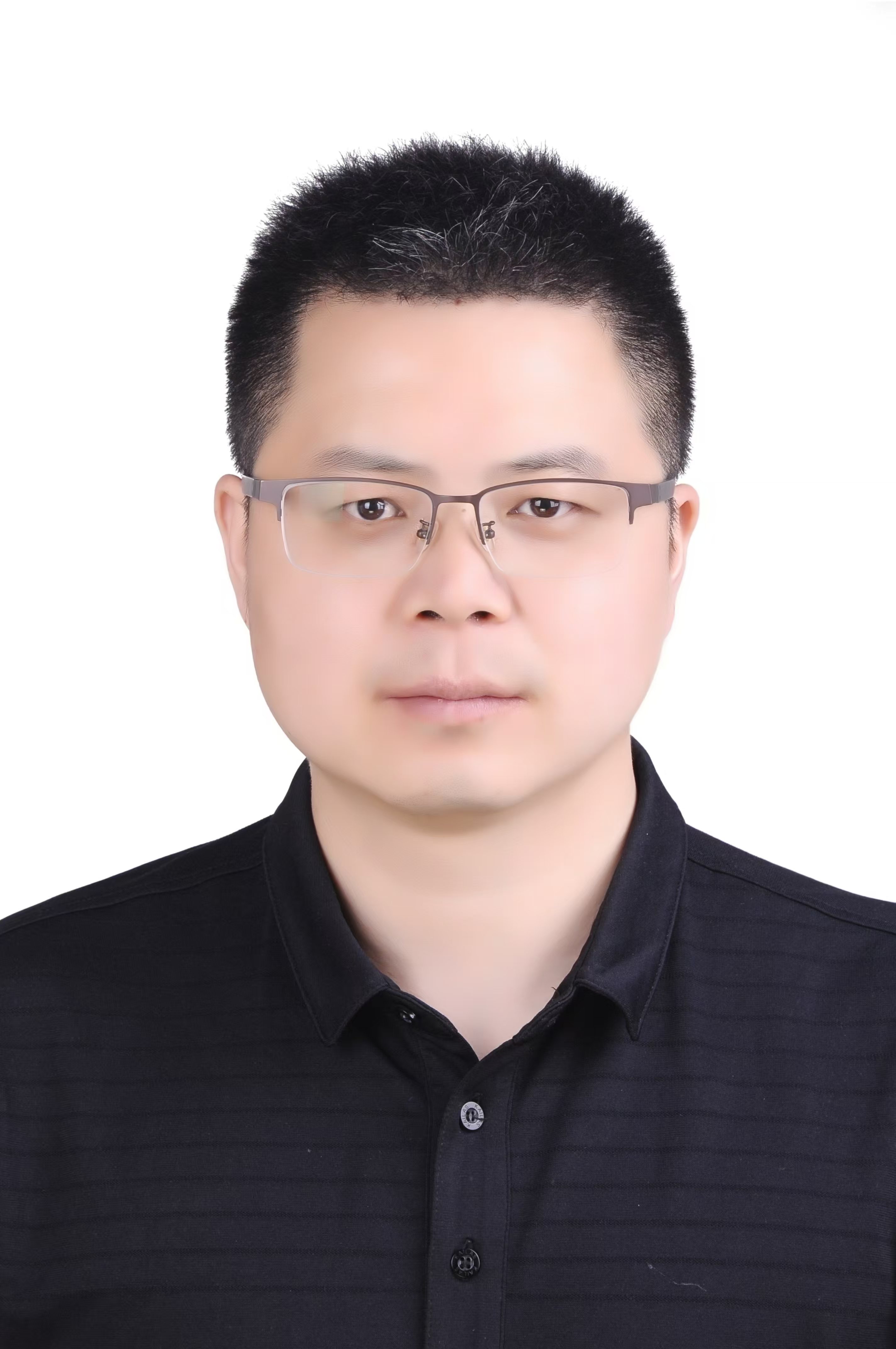}}]{Hao Wang} received his M.S. and Ph.D. degrees from the University of Electronic Science and Technology and the University of Chinese Academy of Sciences in 2008 and 2017, respectively. 

He is currently a research fellow at the Xi'an Institute of Optics and Precision Mechanics, Chinese Academy of Sciences. His research focuses on aerospace visual imaging and image processing.
\end{IEEEbiography}



\begin{IEEEbiography}[{
\includegraphics[width=1.1in,height=1.3in,clip,keepaspectratio]{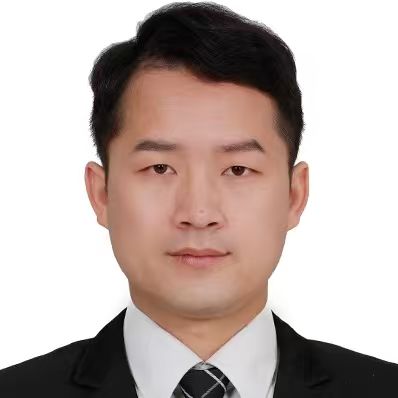}}]{Ming-Ming Cheng} received his PhD degree from Tsinghua University in 2012. Then he did two years research fellow with Prof. Philip Torr in Oxford. 

He is now a professor at Nankai University, leading the Media Computing Lab. His research interests include computer graphics, computer vision, and image processing. He received research awards, including ACM China Rising Star Award, IBM Global SURAward,andCCF-Intel Young Faculty Researcher Program. He is on the editorial boards of IEEE TPAMI/TIP.
\end{IEEEbiography}

\begin{IEEEbiography}[{\includegraphics[width=1in,height=1.25in,clip,keepaspectratio]{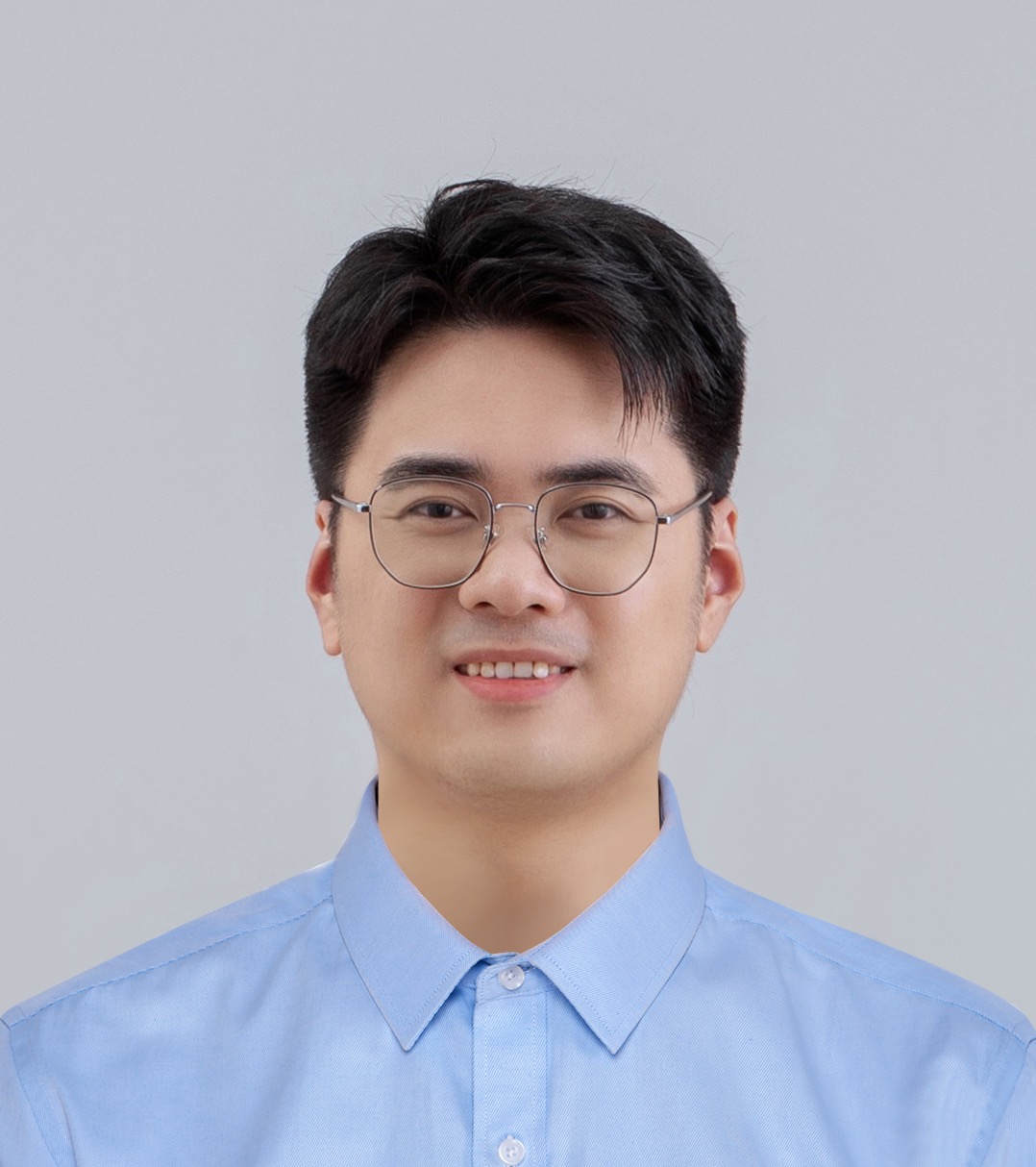}}]{Yimian Dai} 
(Member, IEEE) received the B.E. degree in information engineering and the Ph.D. degree in signal and information processing from Nanjing University of Aeronautics and Astronautics, Nanjing, China, in 2013 and 2020, respectively.

From 2021 to 2024, he was a Postdoctoral Researcher with the School of Computer Science and Engineering, Nanjing University of Science and Technology, Nanjing, China. 
He is currently an Associate Professor with the College of Computer Science, Nankai University, Tianjin, China.
His research interests include computer vision, deep learning, and their applications in remote sensing.
For more information, please visit the link (\href{https://yimian.grokcv.ai/}{https://yimian.grokcv.ai/}).
\end{IEEEbiography}

\end{document}